\documentclass{article}

\PassOptionsToPackage{numbers,compress,sort}{natbib}

\usepackage[main, final]{neurips_2026}

\usepackage[utf8]{inputenc} 
\usepackage[T1]{fontenc}    
\usepackage{hyperref}       
\usepackage{url}            
\usepackage{booktabs}       
\usepackage{amsfonts}       
\usepackage{nicefrac}       
\usepackage{microtype}      
\usepackage{xcolor}         
\usepackage{amsmath}
\usepackage{graphicx}
\usepackage{cleveref}
\usepackage{subcaption}
\usepackage{multirow}

\title{Learnable Spectral Activations}

\author{%
Tamir Shor$^{1}$\thanks{Corresponding author.}
\quad
Or Litany$^{1,3}$
\quad
Alex M. Bronstein$^{1,2}$\\[2pt]
$^{1}$Technion\\
$^{2}$Institute of Science and Technology Austria\\
$^{3}$NVIDIA\\[2pt]
\texttt{tamir.shor@campus.technion.ac.il}
\quad
\texttt{orlitany@cs.technion.ac.il}\\
\texttt{alexander.bronstein@ist.ac.at}
}
\begin{document}

\maketitle

\begin{abstract}
Implicit neural representations (INRs) are shaped by the spectral structure induced by their input encodings and activation functions. Existing methods improve fitting primarily by modifying which frequencies are available to the network, through coordinate encodings or periodic nonlinearities. However, frequency access is not the only bottleneck: signals with localized or spatially varying structure require the network to efficiently compose frequencies into multi-harmonic internal responses. We introduce learnable spectral activations (LSA), which replace fixed neuron-level nonlinearities with a residual truncated Fourier series whose harmonic amplitudes are learned during training. LSA does not expand the asymptotic function class. Instead, it changes the factorization of the representation: linear weights select features while activation coefficients control spectral shaping, and the two are updated by separate gradients. Because the activation output is affine in the coefficients given fixed pre-activations, spectral tuning becomes a more direct subproblem compared to architectures where it is entangled with feature selection. Empirically, this factorization concentrates more target-signal energy in the leading eigenmodes of the neural tangent kernel, consistent with improved optimization behavior. Across audio, image, neural radiance field, and neural acoustic field tasks, LSA also improves reconstruction quality. Code is available at \href{https://github.com/tamirshor7/Learnable-Spectral-Activations}{\textcolor{purple}{https://github.com/tamirshor7/Learnable-Spectral-Activations}}
\end{abstract}

\section{Introduction}
\label{intro}
Implicit neural representations (INRs), or neural fields~\cite{xie2022neural,sitzmann2020implicit}, parameterize signals as continuous functions, mapping coordinates directly to signal values and replacing discrete grids with compact, resolution-agnostic representations that are amenable to querying, differentiation, and optimization. They have driven progress across neural rendering~\cite{mildenhall2021nerf,peng2021neural,muller2022instant}, geometric scene understanding~\cite{park2019deepsdf,mescheder2019occupancy,jiang2020local,williams2022neural,huang2023neural,yang2023emernerf}, acoustic modeling~\cite{lan2024acoustic,luo2022learning}, physical simulation and scientific computing~\cite{zhu2025implicit,raissi2019physics,kadkhodaie2020solving,li2023pac}, sensor simulation~\cite{borts2024radar}, and a growing list of general-purpose use cases~\cite{dupont2022data,shor2025adversarial}. Across all of these, INR behavior is governed by two architectural choices: the input encoding that lifts coordinates into a feature space, and the nonlinear activation through which those features propagate. Together, these choices determine the family of signals the network can fit and how easily it converges to a particular target, making them, in practice, the central design question for INRs.

A productive lens for reading INR design is the structured-dictionary perspective of \citet{yuce2022structured}: an INR represents a signal as a weighted combination of harmonic atoms drawn from a dictionary $\mathcal{H}(\Omega)$ induced by the architecture. The input encoding determines a base set of frequencies $\Omega$, nonlinear activations generate harmonics of those frequencies, and the network learns coefficients that combine the resulting atoms into the target signal. Under this view, designing an INR amounts to choosing and constraining a dictionary, and the inductive bias of an INR is essentially the bias of that dictionary~\cite{tancik2020fourier,jacot2018neural}. Most successful INR methods control this dictionary by adjusting \emph{which} atoms it contains. Fourier features~\cite{tancik2020fourier} prescribe $\Omega$ at the input. SIREN~\cite{sitzmann2020implicit} uses a sinusoidal activation $\sin(\omega_0 u)$ whose frequency scale $\omega_0$ controls the bandwidth the network can resolve. FINER~\cite{liu2024finer,zhu2024finer++} introduces variable-periodic activations whose local frequency depends on the pre-activation magnitude. Meta-learning approaches reshape the dictionary through the initialization~\cite{sitzmann2020metasdf,tancik2021learned}. What unifies all of these approaches is their target: they modify \emph{which} frequencies are available, while leaving the neuron-level nonlinearity that \emph{shapes} those frequencies prescribed by the architecture. This creates a structural limitation that becomes visible on heterogeneous signals. Consider reconstructing an image edge -- a single spatial feature that requires coordinated energy across many harmonics with specific relative amplitudes. In SIREN, each hidden unit produces a response whose harmonic content is rigidly determined by its weights. In FINER, the effective frequency varies with the input, but at any given input each unit still produces a single local oscillatory mode. To reconstruct the edge, both architectures must assemble it from neurons whose spectral content is locked to the same weights that define their input features. The spectral profile of a neuron cannot be adjusted without simultaneously altering the feature it computes.

We introduce \textit{learnable spectral activations} (LSA), which decouple these two roles. At each layer, the fixed activation is replaced by a residual truncated Fourier series whose harmonic amplitudes are learned alongside the network weights. A residual design ensures the activation starts near the identity and progressively adapts a layer-specific harmonic profile during training. This gives each hidden unit a learned, multi-harmonic response over the same pre-activation, without changing the linear weights that select the input feature.

Crucially, consistent with the sinusoidal equivalence perspective of~\citet{morsali2026staf}, LSA does not expand the asymptotic function class: a sufficiently wide periodic INR can represent the same signals. The contribution is a \emph{factorization} change. The linear weights determine which feature each neuron computes while the activation coefficients determine how that feature is spectrally shaped, and the two are updated by different gradients within a jointly optimized parameterization. Because the activation output is affine in the coefficients given fixed pre-activations, spectral tuning is a more direct optimization subproblem than in architectures where spectral control and feature selection are entangled in the same parameters. This factorization has a meaningful effect on training dynamics. The empirical neural tangent kernel (NTK), whose leading eigenfunctions can be interpreted as high-eigenvalue dictionary atoms that gradient descent exploits most efficiently~\cite{jacot2018neural,tancik2020fourier}, concentrates substantially more target-signal energy in its top modes under LSA than under SIREN or FINER (\cref{fig:ntk_energy}).

Experimentally, we evaluate LSA across audio fitting (speech and musical signals), image reconstruction (pixel-domain and gradient-only Poisson reconstruction), neural radiance fields, and neural acoustic fields. Across these controlled activation comparisons, LSA improves reconstruction quality and reduces structured spatial and spectral residuals. Audio experiments also show more stable off-grid behavior, supporting the central conclusion that INR performance depends on how available frequencies are composed inside the network.

\section{Related Work}
\label{sec:related}
This work builds on the structured dictionary perspective of INRs, presented by~\citet{yuce2022structured}. In this view, a dictionary is a collection of elementary functions (\emph{atoms}), whose weighted combination represents a target signal. For INRs, these atoms are induced by the coordinate mapping and activation function. Coordinates $x$ are often first uplifted to an input representation $\gamma(x)$, such as Fourier~\cite{tancik2020fourier} or positional features, to mitigate spectral bias~\citep{rahaman2019spectral}. As shown by~\citet{yuce2022structured}, when $\gamma$ contains sinusoidal components with base frequencies $\Omega$, nonlinear activations generate harmonics, schematically, $\rho(e^{j\omega x})=\sum_k \alpha_k e^{jk\omega x}$. As a result, many INR architectures can be viewed as representing signals as
\begin{equation}
\label{eq:dict}
f(x) = \sum_{\omega' \in \mathcal{H}(\Omega)} c_{\omega'} \sin(\langle \omega', x \rangle + \phi_{\omega'}),
\end{equation}
where $\mathcal{H}(\Omega)$ denotes the harmonic frequencies induced by the architecture and $c_{\omega'}$ are learned coefficients. This perspective connects INR design to the atoms available to the model: input encodings, activations, and initialization schemes determine the dictionary from which the signal is assembled.

Many INR methods can be interpreted as mechanisms for controlling this dictionary. Fourier features~\cite{tancik2020fourier} modify the input mapping directly, using $\gamma(x)=[\sin(2\pi Bx),\cos(2\pi Bx)]$, where $B$ determines the frequencies exposed to the MLP. In random Fourier features, sampling $B_{ij}\sim\mathcal{N}(0,\sigma^2)$ makes $\sigma$ a bandwidth parameter. Larger values expose higher frequencies but can also increase aliasing risk off the training grid~\citep{yuce2022structured}. Positional encodings in neural rendering follow the same principle by prescribing coordinate-frequency bases before the network~\citep{mildenhall2021nerf}. Activation-based INRs instead shape the dictionary inside the network. SIREN~\cite{sitzmann2020implicit} uses hidden layers $z^{(\ell)}=\sin(\omega_0(W^{(\ell)}z^{(\ell-1)}+b^{(\ell)}))$, where $\omega_0$ controls the effective frequency scale. This enables high-fidelity fitting and well-behaved derivatives, since derivatives of sinusoids remain sinusoidal. FINER~\citep{liu2024finer} further tunes spectral bias through a variable-periodic activation $\rho_{\mathrm{FINER}}(u)=\sin((|u|+1)u)$. By changing the bias initialization range, different activation regions with different effective local frequencies are selected. FINER++~\citep{zhu2024finer++} extends this idea to a broader family of variable-periodic functions. SIREN and the FINER family show that activation design is central to INR performance, but remain primarily frequency-centric: they tune which effective frequencies are available or emphasized by a prescribed nonlinear operator. WIRE~\citep{saragadam2023wire} and sinc activations~\citep{saratchandran2024sampling} provide localized Gabor and sampling-theoretic alternatives. Scaling initialization~\citep{yeom2025fast} improves the training of sinusoidal fields through their initial parameter scale. SAPE~\citep{hertz2021sape} and modulated periodic activations~\citep{mehta2021modulated} address related limitations through adaptive input-frequency schedules and latent-code-conditioned modulation, respectively. Multiplicative filter networks~\citep{fathony2020multiplicative} compose learned filters through multiplicative interactions. These mechanisms act on different parts of INR design, including the activation, initialization, input-frequency schedule, conditioning, and network composition.

Recent methods also make the activation or internal basis itself learnable. Among them, STAF~\citep{morsali2026staf} is the closest conceptual baseline to LSA. STAF learns layer-shared sinusoidal mixtures with trainable amplitudes, frequencies, and phases, allowing both the sinusoidal basis and its weighting to adapt during training. LSA adopts a different inductive bias: it fixes an integer-harmonic basis of each pre-activation and learns its relative amplitudes. This choice is motivated by the structured-dictionary analysis of~\citet{yuce2022structured}, where periodic nonlinearities already induce rich harmonic dictionaries and optimization depends not only on available frequency support, but also on how the induced atoms are weighted and aligned with the target. LSA therefore focuses adaptation on the composition of coordinated harmonics within a structured basis rather than on learning the sinusoidal basis itself. The residual form supports initialization near the identity, and layer sharing keeps this adaptation low-dimensional with only $K$ additional coefficients per layer. Fourier KAN~\citep{mehrabian2025fourier}, SL$^2$A~\citep{rezaeian2025sl2a}, and MIRE~\citep{jayasundara2025mire} explore learned Fourier expansions, a Chebyshev activation block within a modulated architecture, and layer-wise activation dictionary matching, respectively. These methods span both activation substitutions and broader architectural changes. We therefore distinguish controlled activation comparisons under matched backbones from broader comparisons with alternative INR formulations.

The dictionary view\citep{yuce2022structured} also connects representation to optimization through the empirical neural tangent kernel (NTK)\citep{jacot2018neural}: leading NTK eigenfunctions can be interpreted as high-eigenvalue atoms that gradient descent learns more rapidly~\citep{jacot2018neural,yuce2022structured}. Thus, improving an INR is not only a matter of expanding frequency support, but also of aligning the induced basis with the target signal. Our work follows this line of reasoning but modifies a different axis of the representation. Rather than further tuning which frequencies are available, LSA learns how a structured harmonic basis is composed within each hidden response, adapting neuron-level spectral shaping while retaining a compact layer-wise parameterization.

\section{Method}
\label{sec:method}
The dictionary view in \Cref{sec:related} suggests that fixed INR activations prescribe the neuron-level waveform family through which hidden features are transformed. We make this operator learnable. At layer \(\ell\), we parameterize the activation as a residual truncated Fourier series:
\begin{equation}
\phi^{(\ell)}(u) = u + \sum_{k=1}^{K} a^{(\ell)}_k \sin(2\pi k u),
\label{eq:method_activation}
\end{equation}
where $u$ denotes a scalar pre-activation and $a_k^{(\ell)}$ are learnable layer-wise coefficients. 
When applied to a vector pre-activation 
$u^{(\ell)} = W^{(\ell)}z^{(\ell-1)} + b^{(\ell)}$, 
$\phi^{(\ell)}$ is applied elementwise with the same coefficients shared across channels. We integrate this activation into a standard INR pipeline. Given coordinates $x$, optionally encoded by $\gamma$, the network computes
\begin{equation}
z^{(0)} = \gamma(x), \qquad
z^{(\ell)} = \phi^{(\ell)}\big(W^{(\ell)} z^{(\ell-1)} + b^{(\ell)}\big),
\end{equation}
for $\ell = 1,\dots,L-1$, followed by a linear output layer. The activation parameters $a^{(\ell)} \in \mathbb{R}^K$ are shared across neurons within each layer and learned jointly with the network weights. They are initialized with small variance and zero mean so that the model starts close to the identity-residual MLP and progressively adapts its nonlinear basis during training. Notably, like SIREN~\citep{sitzmann2020implicit}, LSA provides analytic derivatives through automatic differentiation, since the activation remains a smooth harmonic residual operator. Thus, the model retains the derivative-based advantages of periodic INRs for gradient supervision and physical modeling, while allowing the waveform itself to adapt during training.

Under this formulation, input encodings such as Fourier features determine the \emph{input spectral support}, while LSA controls \emph{internal spectral processing}. The linear weights determine which hidden features are formed, whereas the activation coefficients determine how those features are spectrally shaped. Because the learned operator is applied pointwise to each pre-activation, LSA can adapt the harmonic response of hidden units without changing their underlying feature projections.

The key distinction from prescribed INR activations such as SIREN, WIRE, and the FINER family~\citep{sitzmann2020implicit,saragadam2023wire,liu2024finer,zhu2024finer++} is the factorization of spectral shaping (and, importantly, not asymptotic expressivity). Competitive INR architectures can represent complex waveforms by combining many units, but their neuron-level nonlinear operator remains prescribed. In contrast, LSA in \cref{eq:method_activation} applies multiple harmonic transformations to the same pre-activation $u=w^\top z$ and learns their relative amplitudes, allowing each hidden response to adapt a coordinated multi-harmonic waveform directly rather than relying solely on composition across units with predefined nonlinearities. More generally, a parameterization $u+\sum_k a_k b_k(u)$ with any fixed basis remains affine in the coefficients $a_k$. The harmonic basis specifies the structured inductive bias studied in this work.

\begin{figure}[t]
    \centering
    \begin{subfigure}[t]{0.58\linewidth}
        \centering
        \includegraphics[width=\linewidth]{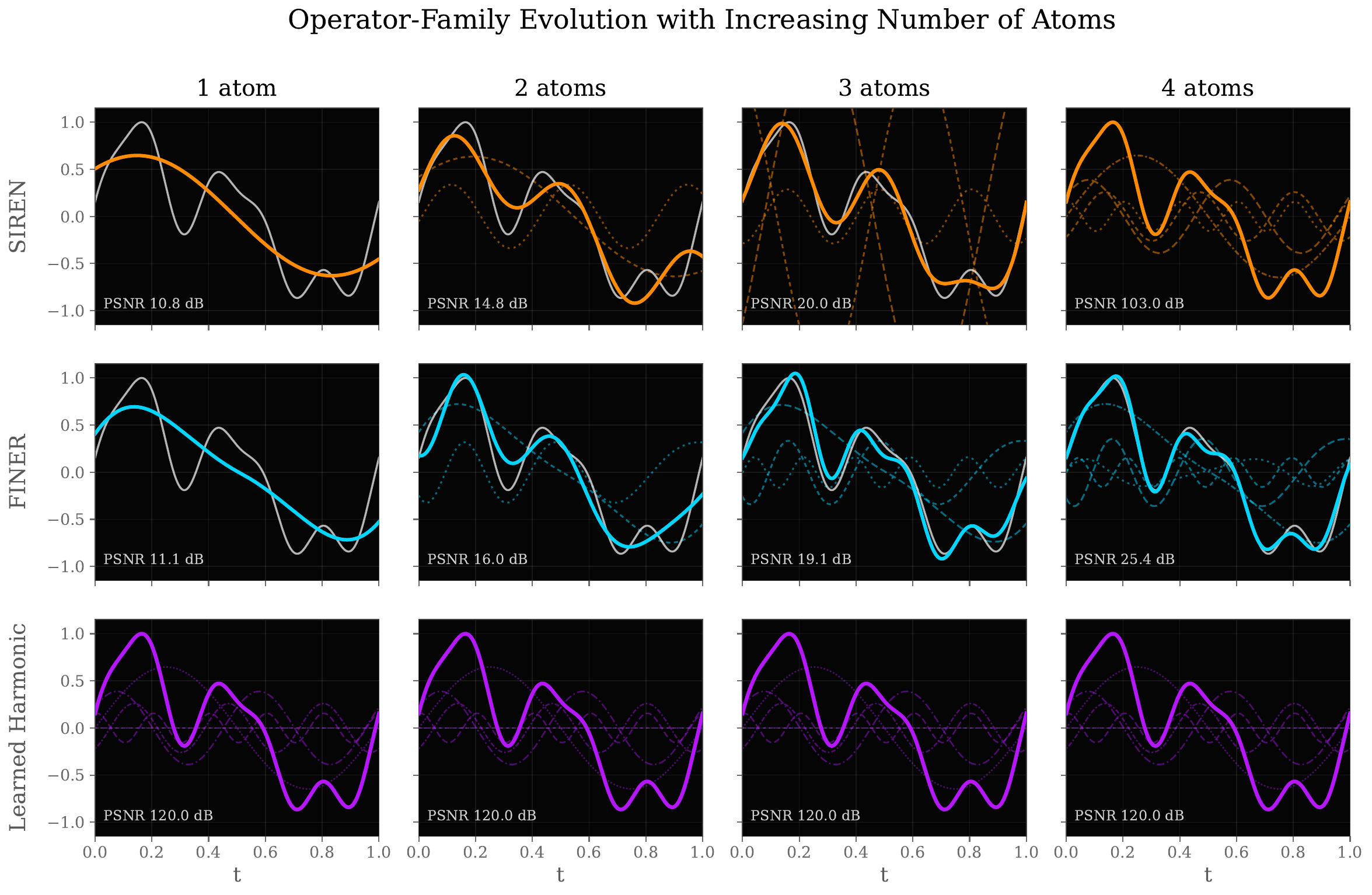}
        \caption{}
        \label{fig:operator_motivation}
    \end{subfigure}
    \hfill
    \begin{subfigure}[t]{0.38\linewidth}
        \centering
        \includegraphics[width=\linewidth]{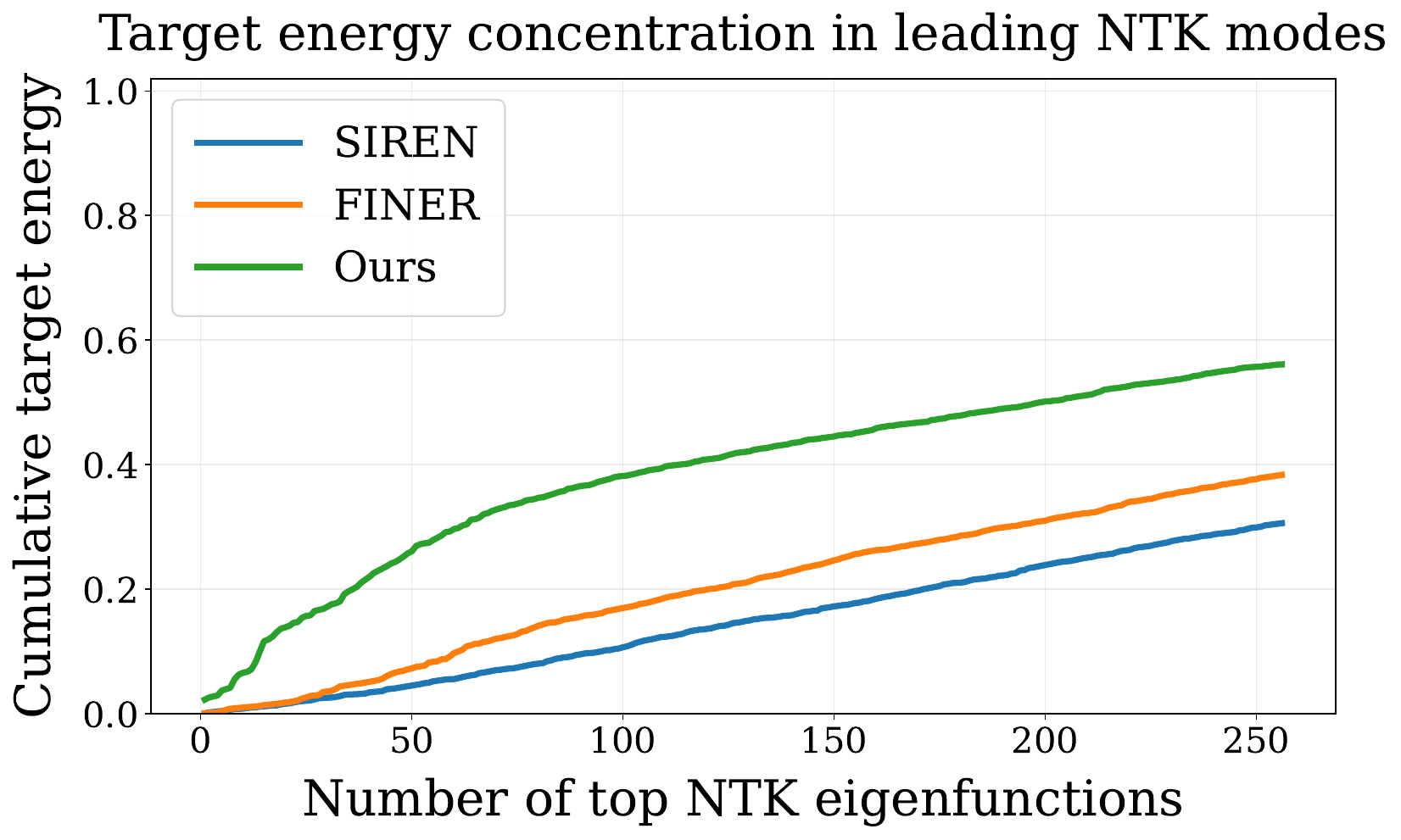}
        \caption{}
        \label{fig:ntk_energy}
    \end{subfigure}
    \caption{ 
    \textbf{(a)} Operator families and parameter efficiency. Dashed curves show individual atoms and solid curves show the fitted signal. SIREN and FINER use prescribed single-unit responses, whereas LSA synthesizes multi-harmonic waveforms and reaches accurate fits with fewer atoms.
    \textbf{(b)} Target energy concentration in leading empirical NTK modes. LSA places more target energy in the top modes than SIREN or FINER, indicating better alignment between the induced dictionary and the target.
    }
    \label{fig:motivation_ntk}
\end{figure}

\Cref{fig:motivation_ntk}a illustrates this distinction on a controlled 1D phantom with a limited atom budget and matched optimization budget. SIREN and FINER improve as more atoms are added, showing that the target is representable by sufficiently many fixed units. LSA reaches the target structure with fewer atoms because each unit can synthesize a harmonic mixture. This experiment illustrates the parameter efficiency of the harmonic factorization.

To characterize these learning directions, let $w$ collect ordinary weights and biases, $a$ collect the harmonic coefficients, and $r=f_{w,a}(X)-y$ be the residual over fixed samples $X$, with output channels flattened. Define $J_w=\partial f(X)/\partial w$ and $\Psi=\partial f(X)/\partial a$. Each column $\psi_{\ell k}$ of $\Psi$ propagates $\sin(2\pi k u^{(\ell)})$ through the remaining layers. For $\mathcal L=\tfrac12\|r\|_2^2$, full-batch gradient flow with positive group rates $\eta_w,\eta_a$ gives the exact finite-width identity
\begin{equation}
-\dot{\mathcal L}=\eta_w\|J_w^\top r\|_2^2+\eta_a\|\Psi^\top r\|_2^2,
\qquad \|\Psi^\top r\|_2^2=\sum_{\ell,k}\langle r,\psi_{\ell k}\rangle^2.
\label{eq:descent}
\end{equation}
Thus coefficient learning follows residual correlations with propagated harmonic features. At a shared initialization, the rate-weighted empirical tangent kernel~\citep{jacot2018neural} satisfies
\begin{equation}
\kappa_{\mathrm{LSA},0}=\kappa_{\mathrm{frozen},0}+\eta_a\Psi_0\Psi_0^\top,
\qquad \kappa_{\mathrm{frozen},0}=\eta_w J_{w,0}J_{w,0}^\top.
\label{eq:kernel_update}
\end{equation}
Here the frozen model holds $a$ at that same initialization. The added positive-semidefinite component contributes $\eta_a\|\Psi_0^\top r_0\|_2^2$ to the initial loss-decay rate. At $a=0$, $\phi(u,0)=u$ and $\partial\phi/\partial a_k=\sin(2\pi ku)$, retaining harmonic tangent directions whenever their downstream propagation is nondegenerate. Frequency and phase derivatives of $A\sin(\omega u+\varphi)$ scale with $A$. Thus the parameterization admits exact identity initialization while retaining active harmonic tangent directions. The optimization benefit is local and conditional on residual alignment. Derivations, the fixed-NTK interpretation, and controlled validations are given in \Cref{app:lsa_theory}.

To further probe this NTK-based optimization view, we analyze the empirical NTK induced by each architecture. 
We eigendecompose the empirical NTK on a fixed set of sampled coordinates and measure how much zero-mean target energy lies in the span of the top-$m$ eigenfunctions. 
These leading eigenfunctions correspond to high-eigenvalue directions that gradient descent learns most efficiently~\citep{yuce2022structured,jacot2018neural}. 
As shown in \Cref{fig:motivation_ntk}b, LSA captures substantially more target energy using the same number of leading modes than SIREN or FINER. 
This suggests that the learned activation reshapes the local optimization geometry so that the target is represented more efficiently by the high-eigenvalue subspace. 
Full definitions and computation details are provided in \Cref{app:ntk_details}.


\section{Results}
\label{sec:results}

We evaluate learned spectral activations across characteristic INR settings. In \Cref{sec:results_audio} we study direct signal fitting on one-dimensional audio, where the target contains transients, harmonic structure, and time-varying envelopes. In \Cref{sec:results_images} we evaluate image fitting, including both pixel-domain reconstruction and a gradient-only Poisson reconstruction task, to test whether the learned activation improves function values and derivative consistency. Next, in \Cref{sec:results_nerf} we move from fitting individual signals to learning neural fields, using novel view synthesis and acoustic-field prediction to test whether LSA improves generalization to unseen spatial queries. In \Cref{app:pe} we further demonstrate that the same spectral parameterization can also be used beyond INR activations, by applying it to positional encodings in a sparse-to-dense reconstruction task.

\paragraph{Baselines.}
Our primary controlled comparisons use SIREN~\citep{sitzmann2020implicit} and FINER~\citep{liu2024finer}, strong prescribed-activation INR baselines that allow us to isolate the effect of replacing the neuron-level nonlinear operator under matched MLP backbones and task settings. We additionally compare with STAF~\citep{morsali2026staf}, the closest learnable-activation baseline to LSA, on both image and audio fitting using independently tuned, task-appropriate configurations. Broader INR comparisons include SL$^2$A\citep{rezaeian2025sl2a}, WIRE\citep{saragadam2023wire}, MFN GaborNet\citep{fathony2020multiplicative}, and fractional Jacobi blocks (fJNB)~\citep{aghaei2024fkan} on compatible tasks. Hyperparameters are selected separately for each method, with protocols reported alongside the comparisons and in \Cref{app:additional_baselines}.

\subsection{Audio Data}
\label{sec:results_audio}
\subsubsection{Audio Signal Fitting}
\label{sec:results_audio_train}

We evaluate our LSA on one-dimensional audio INR fitting. We use LibriSpeech~\cite{panayotov2015librispeech}, a corpus of read English speech derived from LibriVox audiobooks, and NSynth~\cite{engel2017neural}, a dataset of four-second monophonic musical notes from a large collection of instruments. The selected clips are resampled to 48 kHz as detailed in \Cref{app:impaudio}. These two datasets probe complementary regimes: speech contains transient, nonstationary waveform structure, NSynth contains harmonic musical signals with instrument-dependent timbre and temporal envelopes. Because audio contains strong high-frequency and highly nonstationary structure, in the controlled SIREN/FINER/LSA comparison we use Fourier features~\citep{tancik2020fourier} at the input layer and include raw coordinates as the $\sigma=0$ setting. For STAF, we also evaluated matched Fourier-feature settings, but its native raw-coordinate configuration performed best. We sweep the Fourier bandwidth $\sigma$ and tune it independently for each method. The comparison therefore examines how each activation processes the same exposed frequency support at each bandwidth. Full implementation details are available in \Cref{app:impaudio}.

\begin{figure}[t]
\centering
\includegraphics[width=0.8\linewidth]{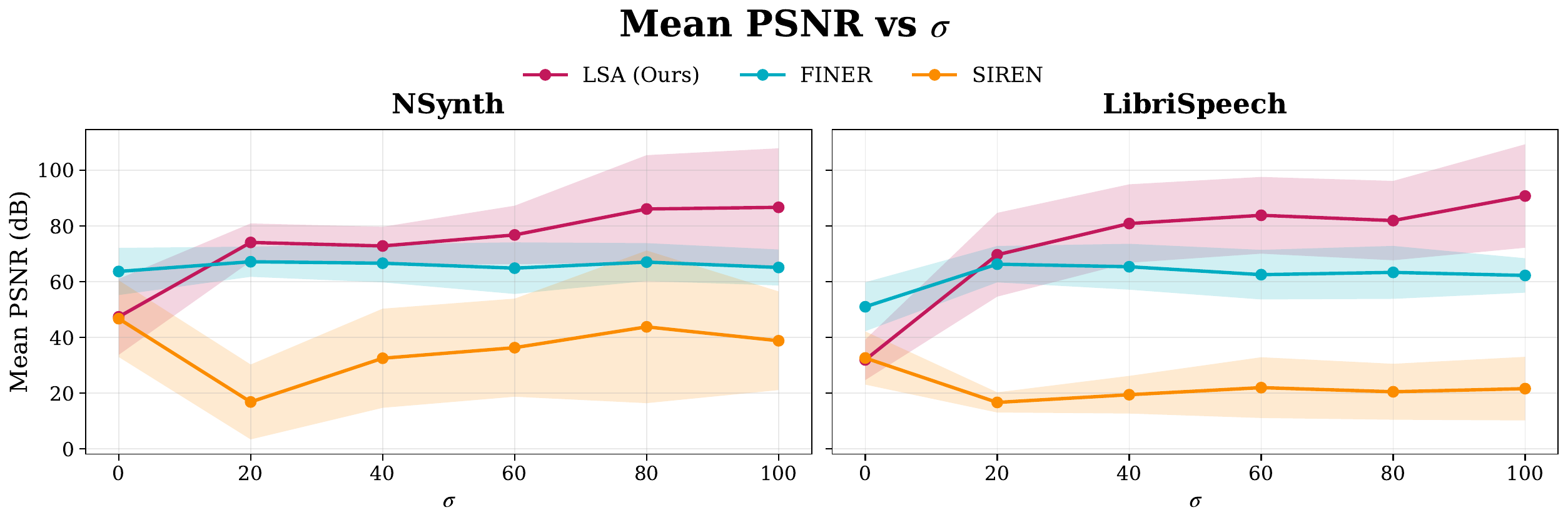}
\caption{
SIREN/FINER/LSA comparison of on-grid audio fitting across Fourier-feature bandwidths $\sigma$. LSA achieves the strongest performance at moderate and high bandwidths on both speech and musical-note signals, indicating that learned spectral activations improve the internal processing of Fourier-feature harmonics.
}
\label{fig:audio_quant}
\end{figure}

\Cref{fig:audio_quant} reports on-grid PSNR as a function of the Fourier-feature bandwidth $\sigma$. With raw coordinates ($\sigma=0$), FINER achieves slightly higher PSNR on both LibriSpeech and NSynth. This is consistent with our formulation: LSA does not expose a different input basis, but rather gives each layer a learnable mechanism for reshaping the spectral components made available to the network. Following Tancik et al.~\citep{tancik2020fourier}, we use $\gamma_\sigma(t)=[t,\sin(2\pi Bt),\cos(2\pi Bt)]$ with entries of $B$ drawn from $\mathcal{N}(0,\sigma^2)$ and fixed across methods for a given setting. As $\sigma$ increases, the advantage of LSA becomes clear. On LibriSpeech, LSA improves sharply from the no-FF setting and achieves the best mean performance at positive bandwidths in the sweep, reaching the highest PSNR at large bandwidths. FINER is competitive at low bandwidth but saturates around the mid-60 dB range, while SIREN remains substantially lower and more variable. This supports the view that INR fitting, especially for challenging high-frequency signals (such as speech or audio), benefits not only from exposing selected frequencies at the input, but from learning how hidden responses should synthesize transient and nonstationary waveform structure. On NSynth, the same trend holds. FINER is strong and relatively stable, reflecting the more global harmonic structure of musical notes, but LSA overtakes it once Fourier-feature support is available and maintains the best performance at moderate and high $\sigma$. SIREN is again considerably more sensitive to bandwidth, indicating that under matched Fourier-feature settings, the advantage of LSA comes from processing the exposed frequency support more effectively rather than from a different input basis.

Comparisons with STAF, SL$^2$A, WIRE, and fJNB, including raw-coordinate and Fourier-feature settings, are in \Cref{app:baseline_audio}. Specifically we show that LSA also outperforms STAF in our tested setting - under the same clip manifests and 5,000-step budget, LSA reaches $86.72$/$90.78$ dB on NSynth/LibriSpeech, compared with $68.79$/$68.40$ dB for the strongest tested STAF configuration.

\subsubsection{Continuous Approximation}
\label{sec:continuous_approx}

In \Cref{sec:results_audio_train} we evaluate standard INR fitting on discrete training grids. We now examine the quality of the represented \emph{continuous} signal between sampled coordinates. This distinction is both practical and theoretical: mitigating spectral bias \cite{rahaman2019spectral} via high-frequency injection (e.g., Fourier features) improves fitting on training coordinates, but can introduce aliasing, where multiple high-frequency functions agree on sampled points while diverging between them \cite{yuce2022structured}. 

\begin{figure}[t]
\centering
\includegraphics[width=0.8\linewidth]{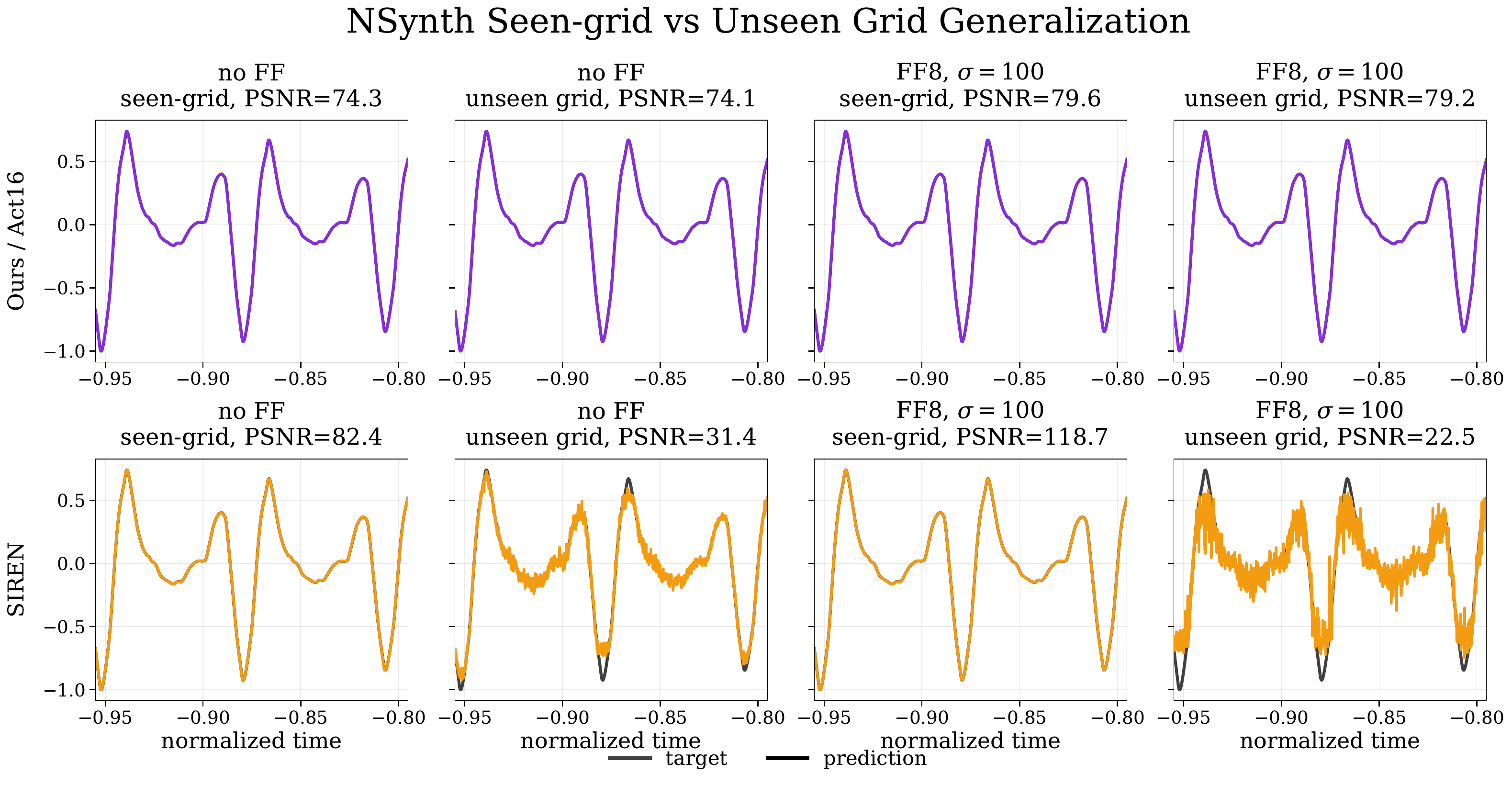}
\caption{On-grid vs. off-grid reconstruction under increasing Fourier-feature bandwidth. SIREN exhibits a large on-grid/off-grid gap already at $\sigma=0$ and severe interpolation artifacts for $\sigma=100$. Our method remains stable under inference of interpolated, unseen grid points.}
\label{fig:nsynth_seen_unseen_mini}
\end{figure}

To isolate this effect, we train on a fixed grid of 16k samples and evaluate both \emph{on-grid} and an \emph{unseen midpoint grid}, while varying high-frequency injection (governed by Fourier-feature bandwidth $\sigma$). 
Extending beyond the common INR use-case of memorizing training samples, in this section we evaluate how well INR performs \emph{continuous} reconstructions, necessitating interpolation between samples seen during INR fitting. We vary Fourier-feature bandwidth via $\sigma$, following standard practice for mitigating spectral bias. Figure~\ref{fig:nsynth_seen_unseen_mini} compares prediction on seen- and unseen-samples. For SIREN, increasing bandwidth ($\sigma$) improves on-grid PSNR but leads to severe degradation off-grid. This indicates that the model fits sampled points using high-frequency components that do not define a stable continuous function. Full version in Appendix (Fig.~\ref{fig:nsynth_seen_unseen_full}).
In contrast, our method maintains consistent performance across seen and unseen grids, without the same collapse at high bandwidth. From the structured-dictionary perspective~\cite{yuce2022structured}, increasing $\sigma$ broadens the base frequency support exposed by the input encoding and, consequently, the higher-frequency content available in the induced dictionary. While this alleviates spectral bias, it also introduces ambiguity: multiple high-frequency functions can agree on sampled points while differing between them, leading to aliasing \cite{yuce2022structured}. This effect is visible for SIREN, where high-bandwidth inputs enable accurate interpolation at training coordinates but unstable behavior off-grid.
We include a similar comparison with FINER in \Cref{fig:nsynth_seen_unseen_full}. FINER similarly expands the supported frequency set via activation design\cite{liu2024finer}, but does not inherently resolve this ambiguity when external bandwidth is large.

\subsection{Image Data}
\label{sec:results_images}

We evaluate learned spectral activations on image fitting using the Kodak dataset~\cite{KodakDataset}, consisting of 24 true-color natural images of resolution $768 \times 512$. Each image is represented as a coordinate-based function using an MLP, without any input-level uplifting mechanisms applied. We compare against SIREN, FINER, and STAF. The matched-budget and extended-budget STAF protocols are distinguished in \Cref{app:baseline_images}.

\begin{figure}[t]
\centering
\includegraphics[width=\linewidth]{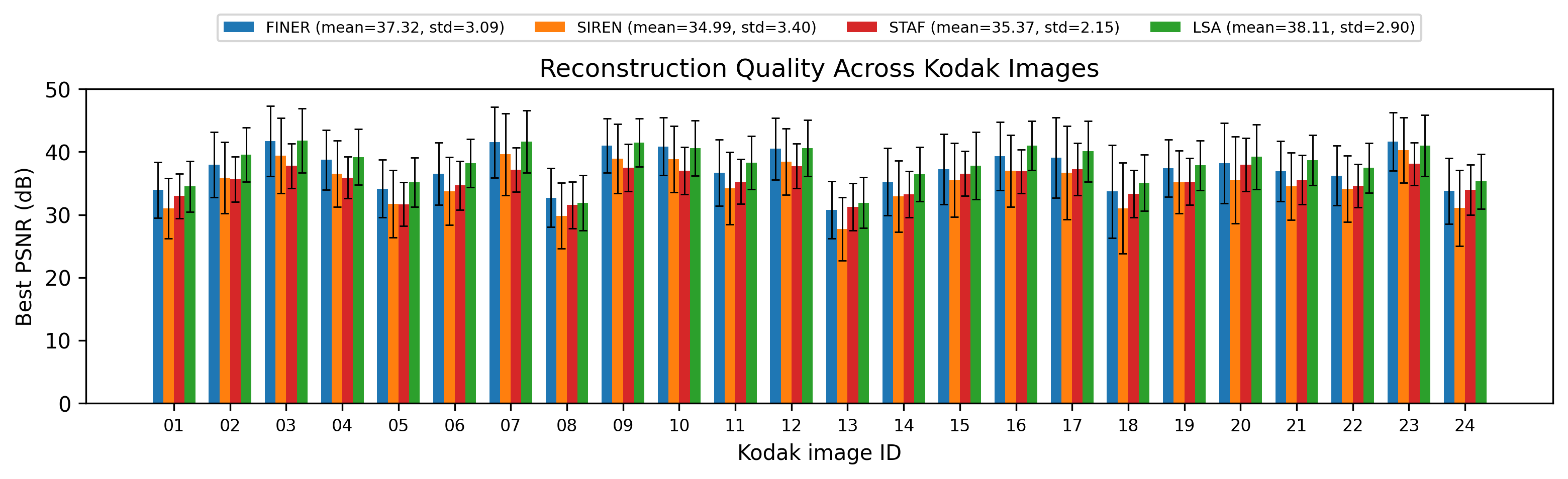}
\caption{
Per-image Kodak fitting comparison under fixed configurations. Per-image PSNR is shown for LSA, FINER, SIREN, and STAF under 5,000-update fits.
}
\label{fig:kodak_bar}
\end{figure}

\Cref{fig:kodak_bar} compares fixed configurations across all 24 Kodak images. LSA obtains a mean PSNR of 38.110 dB, compared with 37.317 dB for FINER, 35.369 dB for matched-budget global STAF, and 34.990 dB for SIREN. LSA exceeds FINER on 21/24 images and SIREN on all 24. Among the competing matched-backbone baselines, FINER achieves the highest mean PSNR. The full-set means include the six calibration images used to select the LSA and baseline configurations. Selection-independent comparisons use the other 18 images, as described in \Cref{app:baseline_images}. With per-image tuning, LSA reaches $38.308$ dB and STAF $35.425$ dB, using three and 56 candidate configurations per image, respectively. LSA wins on all 24 images, with a paired mean gain of 2.883 dB and a 95\% bootstrap interval of $[2.439,3.295]$ dB.

We compare to broader INR methods in \Cref{app:baseline_images}.
On the 18-image holdout, LSA reaches $38.045$ dB with 199,043 trainable parameters, while full SL$^2$A reaches $37.857$ dB with 330,243. LSA has a $0.188$ dB higher descriptive mean while using $\sim40\%$ fewer parameters. At a nearly matched 199,171 parameters, SL$^2$A reaches $35.910$ dB, giving LSA a 2.135 dB mean advantage with 18/18 images where LSA outperforms SL$^2$A. We refer the reader to \Cref{app:baseline_images} for complete protocols and comparisons to WIRE\citep{saragadam2023wire},MFN\citep{fathony2020multiplicative}, and fJNB\citep{aghaei2024fkan}.

\begin{figure}[t]
    \centering
    \includegraphics[width=0.82\linewidth]{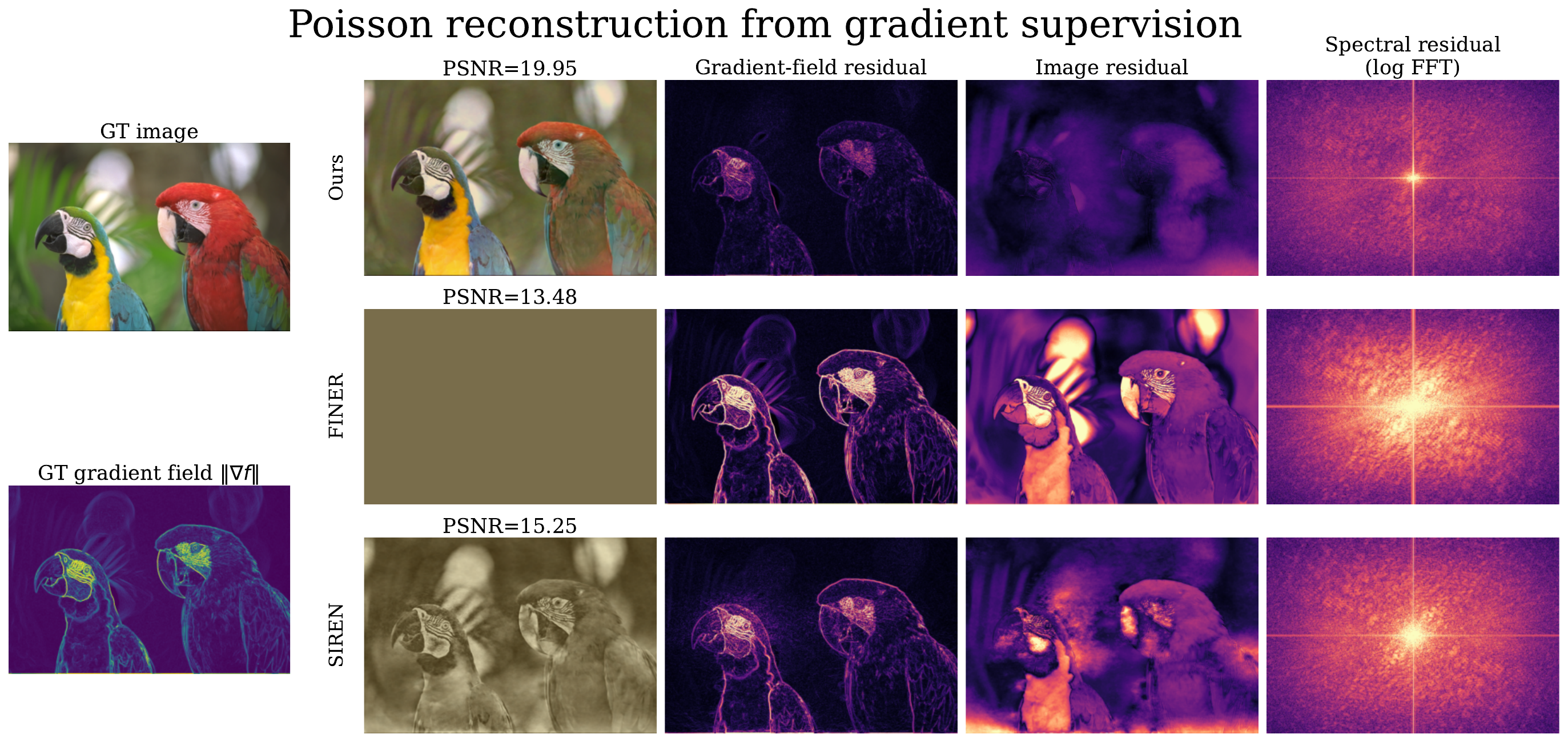}
    \caption{Poisson reconstruction from gradient-only supervision. Our method produces more consistent gradient fields and integrates them into a coherent image, while FINER and SIREN exhibit structured residuals and global reconstruction errors.}
    \label{fig:poisson_grad}
\end{figure}

\paragraph{Poisson Reconstruction -}
To evaluate derivative modeling, we consider a Poisson-style reconstruction task from image gradients. Given a target image $f_{\mathrm{GT}}$, we compute its spatial gradients and train each INR using only the derivative loss $\|\partial_x f_\theta - \partial_x f_{\mathrm{GT}}\|_2^2 + \|\partial_y f_\theta - \partial_y f_{\mathrm{GT}}\|_2^2$, without any direct pixel supervision. Since gradients determine the image only up to an additive constant, we align the mean of each reconstructed RGB channel to the target after training. This setting isolates the ability of the model to learn a derivative field that is both accurate and globally integrable. As shown in \Cref{fig:poisson_grad}, FINER produces highly structured gradient residuals and fails to integrate them into a coherent image, resulting in large low-frequency distortions. SIREN achieves better local structure, but exhibits global drift and inconsistent reconstruction. Our method yields lower gradient-field residuals and a more coherent reconstruction, with reduced structured artifacts in the spectral domain. These results indicate that the learned spectral activation improves the consistency between local derivatives and global signal reconstruction.

\begin{figure}[t]
\centering
\includegraphics[width=\linewidth]{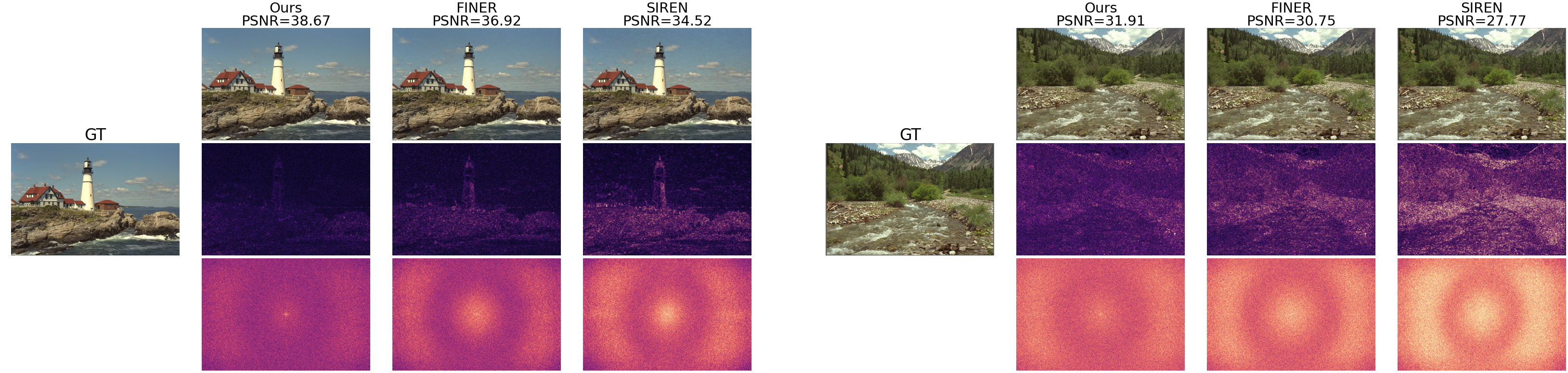}
\caption{
Qualitative Kodak reconstructions for illustrative examples kodim21 (left) and kodim13 (right), selected for qualitative clarity. LSA, FINER, and SIREN use the fixed configurations in \Cref{fig:kodak_bar}. We show reconstructions, spatial residuals, and spectral residuals, with a common display scale for each residual type across both images and all methods. Full-dataset comparisons are in \Cref{fig:kodak_bar} and \Cref{app:baseline_images}.
}
\label{fig:kodak_qual}
\end{figure}

\Cref{fig:kodak_qual} shows illustrative reconstructions and residuals for kodim21 and kodim13. In these examples, LSA reduces structured errors around edges and textures, with corresponding reductions in spectral residuals. The examples were selected for qualitative clarity. Dataset-wide conclusions follow from \Cref{fig:kodak_bar} and the quantitative tables. Natural images contain edges, textures, and repeated fine structure that can be expressed through coordinated harmonic mixtures. \Cref{eq:method_activation} allows each hidden response to adapt such harmonic composition directly. These observations are consistent with the harmonic-composition interpretation in \Cref{fig:operator_motivation}.

The convergence curves in \Cref{app:image_res} show that LSA separates from SIREN early, while its advantage over FINER emerges progressively. LSA leads FINER on 5/24 images at step 500, 16/24 at step 1,000, and 21/24 at step 2,000, finishing ahead on 21/24 at step 5,000.

A six-image ablation improves from 30.45 dB at $K=4$ to 39.50 dB at $K=32$, with no further improvement at $K=64$. Freezing the near-identity coefficients reduces held-out Kodak PSNR from 38.05 to 19.91 dB, showing the contribution of coefficient learning. Full results are in \Cref{app:lsa_ablations}.

\subsection{Neural Fields}
\label{sec:results_nerf}

We next evaluate LSA in neural-field settings, where the objective is to learn a continuous field that generalizes to unseen spatial queries. 

\paragraph{NeRFs.}
For visual neural fields, we evaluate novel view synthesis on the Blender dataset~\citep{mildenhall2021nerf}. The backbone INR model in all experiments is an InstantNGP~\citep{muller2022instant}-style neural field. We test whether the proposed activation improves internal spectral processing beyond isolated coordinate-to-value regression, and whether it supports improved neural-field representation in a setting where the task is to generalize to unseen views. \Cref{tab:nerf_blender_full} reports results across eight object scenes. Learned spectral activations achieve the best PSNR on five scenes and the best SSIM on seven scenes, while matching or improving the strongest LPIPS score on most scenes. The gains are moderate compared to the audio and image fitting settings, but they are consistent across complementary metrics: our method improves perceptual quality and structural similarity even when PSNR differences are small. This suggests that the benefit of LSA is not restricted to one-dimensional or two-dimensional fitting, but also appears when the representation is optimized through differentiable rendering. Qualitative results are in \Cref{app:nf}. Additional fJNB results and published WIRE results under the reduced-view Blender protocol are reported in \Cref{app:baseline_nerf}.

\begin{table}[t]
\centering
\caption{Controlled activation comparison on novel view synthesis, over Blender~\cite{mildenhall2021nerf} scenes.}
\label{tab:nerf_blender_full}
\resizebox{0.82\linewidth}{!}{
\begin{tabular}{llcccccccc}
\toprule
Metric & Method & Chair & Drums & Ficus & Hotdog & Lego & Materials & Mic & Ship \\
\midrule
\multirow{3}{*}{PSNR $\uparrow$}
& SIREN~\citep{sitzmann2020implicit} & 33.48 & 24.89 & 28.22 & 32.85 & 29.60 & 27.13 & 33.28 & 22.48 \\
& FINER~\citep{liu2024finer} & 33.90 & \textbf{24.90} & 28.70 & 33.05 & \textbf{30.04} & 27.05 & \textbf{33.96} & 22.47 \\
& Ours  & \textbf{34.18} & 24.79 & \textbf{28.85} & \textbf{33.36} & 29.39 & \textbf{27.22} & 33.91 & \textbf{22.60} \\
\midrule
\multirow{3}{*}{SSIM $\uparrow$}
& SIREN~\citep{sitzmann2020implicit} & 0.973 & 0.912 & 0.959 & 0.960 & 0.948 & 0.932 & 0.979 & 0.797 \\
& FINER~\citep{liu2024finer} & 0.973 & 0.911 & 0.958 & 0.959 & \textbf{0.951} & 0.928 & 0.981 & 0.792 \\
& Ours  & \textbf{0.977} & \textbf{0.922} & \textbf{0.967} & \textbf{0.967} & 0.950 & \textbf{0.936} & \textbf{0.983} & \textbf{0.803} \\
\midrule
\multirow{3}{*}{LPIPS $\downarrow$}
& SIREN~\citep{sitzmann2020implicit} & 0.017 & 0.051 & 0.027 & 0.032 & 0.030 & \textbf{0.028} & 0.014 & 0.116 \\
& FINER~\citep{liu2024finer} & 0.015 & 0.052 & 0.024 & 0.033 & \textbf{0.025} & 0.032 & \textbf{0.010} & 0.108 \\
& Ours  & \textbf{0.013} & \textbf{0.050} & \textbf{0.021} & \textbf{0.027} & 0.026 & 0.029 & \textbf{0.010} & \textbf{0.101} \\
\bottomrule
\end{tabular}
}
\end{table}

\paragraph{Neural-Acoustic Fields.}
We then test the same principle in acoustic-field modeling. Following Neural Acoustic Fields (NAF)~\citep{luo2022learning}, the task is to represent spatial sound propagation by mapping listener-emitter configurations to room impulse responses. Unlike the audio-fitting experiments in \cref{sec:results_audio_train}, where the model is asked to fit one waveform in isolation, here it must learn a spatially varying field whose responses change with listener position, emitter position, room geometry, occlusion, and reflections. This makes room-impulse-response prediction a complementary field-generalization test to NeRF. In both cases, the network represents a continuous physical field, but the forward structure differs: NeRF predicts visual radiance accumulated along rays, while the acoustic field predicts time-frequency impulse-response structure over spatial acoustic queries.

\begin{figure}[t]
    \centering
    \includegraphics[width=\linewidth]{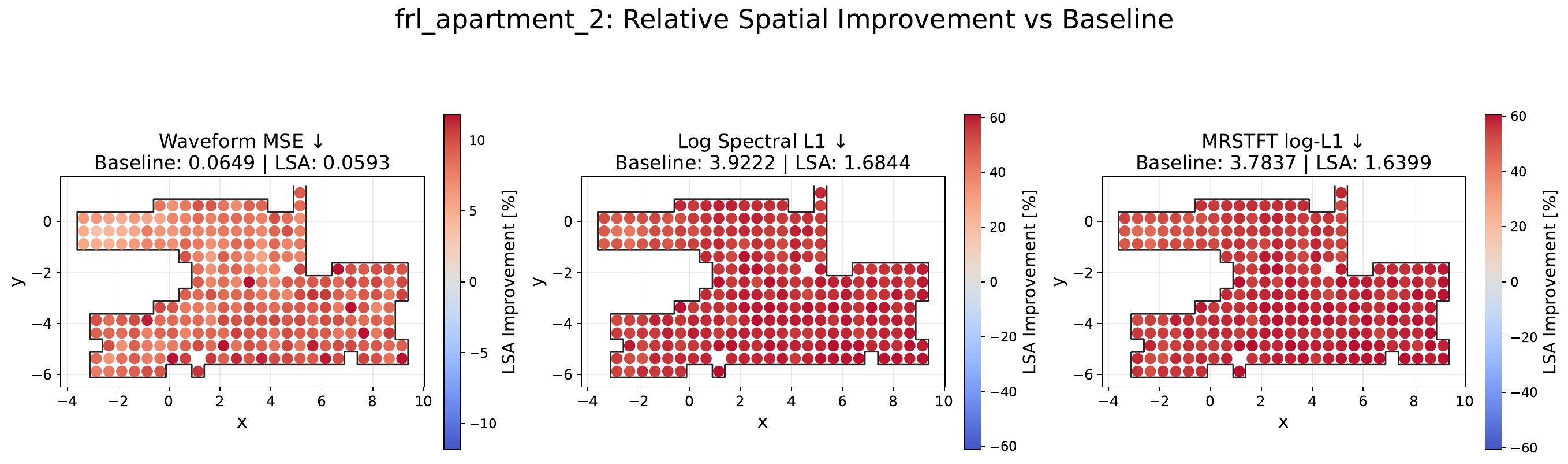}
    \caption{Spatial improvement of LSA over the NAF baseline. Each point corresponds to a spatial query in the scene, and color indicates relative improvement. LSA yields broad spatial gains, especially for log-spectral and multi-resolution STFT metrics, across the queried locations.}
    \label{fig:naf_spatial_improvement}
\end{figure}

We keep the NAF architecture and training protocol fixed and replace its baseline activation with LSA using $K=8$ harmonics. We evaluate on benchmark scenes used in NAF~\cite{luo2022learning}. Full results across the benchmark are reported in \Cref{app:nf}. \Cref{fig:naf_spatial_improvement} shows the spatial improvement in common acoustic reconstruction metrics after replacing standard original NAF activations with LSA. Replacing the baseline activation with LSA yields consistent gains over the original NAF architecture, especially on spectral and time-frequency metrics. 
Across the floor plan, LSA improves the predicted acoustic field, with particularly strong and spatially consistent gains in log-spectral and multi-resolution STFT errors. This matters because acoustic fields are highly heterogeneous: nearby walls, openings, occlusions, and reflections induce different impulse-response structures at different positions. As in NeRF, this requires composing spectral components into spatially varying field responses that remain consistent at unseen queries.
Implementation details and metric definitions are in \Cref{app:naf}

Taken together, the gains across NeRF and NAF settings indicate that beyond advantage in signal-fitting, LSAs also provide a more effective internal parameterization for continuous neural fields, where localized spectral structure must be composed consistently across space.

\section{Conclusion}
\label{sec:conclusion}

In this work we presented learnable spectral activations (LSA) - a simple activation-level parameterization for INRs. Our formulation makes the neuron-level waveform operator trainable by representing it as a residual harmonic expansion. This changes the granularity at which spectral structure is modeled. Hidden units can directly adapt their harmonic response, while the network weights continue to determine the features on which these responses operate. The resulting formulation provides a practical mechanism for learning internal spectral bases that are better aligned with heterogeneous and nonstationary signals. 
Across controlled activation comparisons in audio, images, and neural fields, LSA improves reconstruction quality and reduces structured spatial and spectral residuals. Audio experiments also show more stable off-grid behavior. Our results support the role of learned harmonic composition as a structured inductive bias for INR fitting and spatial-query prediction.

Alongside positive results, a current limitation of our method is increased computational overhead. Although LSA adds very few trainable parameters, evaluating multiple harmonic terms increases activation runtime and memory. In a representative full-grid audio setting, LSA adds 64 parameters but incurs roughly \(7\times\) training-time and \(9\times\) peak-memory overhead relative to FINER, with similar inference overheads. Full measurements are reported in \Cref{app:resource_analysis}. More efficient fused or checkpointed implementations may reduce this cost. Future work should study more efficient and expressive variants of LSA, including neuron-specific, channel-grouped, or input-conditioned harmonic coefficients, as well as regularization of the learned spectra. Overall, our results suggest that learning the internal spectral operators of coordinate networks is a promising route toward more adaptive implicit representations.

\section*{Acknowledgements and Disclosures}

Or Litany acknowledges support from the Israel Science Foundation (grant 624/25) and the Azrieli Foundation Early Career Faculty Fellowship. The authors gratefully acknowledge this support. This research was supported by the Council for Higher Education in Israel under the Moonshot Project. The authors have no competing financial interests to report.

\begingroup
\small
\bibliographystyle{unsrtnat}
\bibliography{bib}
\endgroup

\appendix

\section{Implementation Details}
\label{app:repro}
In this section we elaborate on implementation details across main experiments from \Cref{sec:results}. All experiments are conducted on a single NVIDIA RTX A6000 GPU unless otherwise stated.
\subsection{Audio Data}
\label{app:impaudio}
\label{app:audio_fitting}

\paragraph{Datasets and preprocessing.}
We evaluate audio INR fitting on fixed subsets of LibriSpeech and NSynth. For
LibriSpeech, we use the fixed manifest \texttt{pilot40\_manifest.csv}, containing
40 one-second clips. For NSynth, we use the fixed manifest
\texttt{pilot10\_convergence\_manifest.csv}, containing 10 one-second clips.
All reported quantitative curves use these fixed clip sets. The manifests are
held fixed across methods and hyperparameter sweeps. Audio is resampled to
48 kHz and each training signal is represented by \(N=48{,}000\) samples. We
convert each clip to a single waveform channel and use the waveform amplitudes
on the decoded floating-point audio scale. We do not apply per-method
renormalization. Coordinates are normalized to \(t \in [-1,1]\), and each model
learns a coordinate-to-amplitude function \(f_\theta(t)\in\mathbb{R}\).

\paragraph{Fourier features.}
For coordinate input \(t\in\mathbb{R}\), we use Gaussian Fourier features
\[
    \gamma_{\sigma}(t)
    =
    \left[
    t,\,
    \sin(2\pi t B^\top),\,
    \cos(2\pi t B^\top)
    \right],
    \qquad
    B_i \sim \mathcal{N}(0,\sigma^2),
\]
where \(B\in\mathbb{R}^{M\times 1}\) and \(M\) is the number of Fourier-feature
bands. Unless otherwise stated, \(M=16\). The same Fourier-feature width and the
same sampled matrix \(B\) are used across methods for a given seed and
\(\sigma\). The setting \(\sigma=0\), equivalently \(M=0\), denotes raw
coordinates without Fourier features. We sweep \(\sigma\) and report both the
full sweep and the best setting per method. In the main audio results, the
selected settings are \(\sigma=100\) for LSA with FF16 and \(\sigma=20\) for
FINER with FF16. SIREN is also evaluated under the same matched FF regime when
used in compute comparisons.

\paragraph{Models.}
All audio fitting models are coordinate MLPs with scalar output. Unless
otherwise stated, models use hidden width 256 and depth 4. SIREN uses sinusoidal
activations with \(\omega_0=30\) for both the first and hidden layers. FINER uses
the authors' variable-periodic activation,
\[
    \phi_{\mathrm{FINER}}(u)
    =
    \sin\!\left(\omega_0 (|u|+1)u\right),
\]
with \(\omega_0=30\). We use the FINER-style initialization path used in our
experiments: the first-layer bias scale is set by \(f_{\mathrm{bs}}\), and the
reported audio runs use \(f_{\mathrm{bs}}=20\). Hidden-layer weights and biases
follow the implementation's hidden-layer initialization rule. LSA uses the
residual harmonic activation from \Cref{eq:method_activation},
\[
    \phi_{\mathrm{LSA}}(u)
    =
    u + \sum_{k=1}^{K} a_k \sin(2\pi k u),
\]
including the residual identity term exactly as written. The harmonic
coefficients \(a_k\) are shared within each activation layer, so LSA adds only
\(K\) scalar parameters per activation layer. The main audio configuration uses
\(K=16\). Coefficients are initialized from a zero-mean Gaussian with standard
deviation \(0.01\).

\begin{table}[t]
\centering
\caption{
Audio fitting hyperparameters. FF16 denotes \(M=16\) Gaussian Fourier-feature
bands. The selected \(\sigma\) values are those used for the main audio fitting
results. Full \(\sigma\)-sweep curves are reported separately.
}
\label{tab:audio_hyperparams}
\resizebox{\linewidth}{!}{
\begin{tabular}{ll}
\toprule
Quantity & Setting \\
\midrule
Datasets & LibriSpeech fixed 40-clip manifest. NSynth fixed 10-clip manifest \\
Sampling rate & 48 kHz \\
Samples per clip & \(N=48{,}000\) samples, one-second window \\
Channels & Mono waveform fitting \\
Coordinate range & \(t\in[-1,1]\) \\
Output dimension & 1 waveform amplitude \\
Loss & Waveform MSE \\
Optimizer & Adam \\
Training steps & 5{,}000 \\
Batching & Full grid unless explicitly marked as minibatch. Minibatch runs use 4{,}096 coordinates \\
Hidden width & 256 \\
Depth & 4 hidden activation layers unless stated otherwise \\
Fourier features & \(\gamma_\sigma(t)=[t,\sin(2\pi tB^\top),\cos(2\pi tB^\top)]\) \\
FF width & \(M=16\) bands for FF16 \\
Raw-coordinate setting & \(M=0\), reported as no-FF or \(\sigma=0\) \\
SIREN & \(\omega_0=30\) \\
FINER & \(\omega_0=30\), first-layer bias scale \(f_{\mathrm{bs}}=20\) in audio runs \\
LSA & \(K=16\), residual identity included, layer-shared coefficients \\
LSA coefficient init. & \(a_k\sim\mathcal{N}(0,0.01^2)\) \\
Checkpoint selection & Best observed PSNR over the training trajectory \\
Metric & \(\mathrm{PSNR}=-10\log_{10}(\mathrm{MSE})\), dynamic range \(1\) \\
Curve aggregation & Mean across clips. Shaded regions show standard deviation across clips \\
\bottomrule
\end{tabular}
}
\end{table}

\paragraph{Training and model selection.}
All methods are trained to minimize waveform mean-squared error over coordinate
samples. Unless a minibatch setting is explicitly specified, training uses the
full \(48{,}000\)-sample grid. We use Adam and train for 5,000 optimization
steps. Reported PSNR values use the best checkpoint observed along the training
trajectory rather than the final step. For all method comparisons, the clip
manifests, coordinate grids, Fourier-feature width, and Fourier-feature samples
are held fixed across methods for the corresponding seed and \(\sigma\).

\paragraph{Metrics.}
For a target waveform \(y\) and prediction \(\hat{y}\), we compute
\[
    \mathrm{MSE}
    =
    \frac{1}{N}\sum_{i=1}^{N}(\hat{y}_i-y_i)^2,
    \qquad
    \mathrm{PSNR}
    =
    -10\log_{10}(\mathrm{MSE}),
\]
using unit dynamic range. Dataset-level values are averages over the fixed clip
sets. In \Cref{fig:audio_quant}, each curve reports the mean over clips and the shaded region
denotes the standard deviation across clips.

\paragraph{Continuous approximation.}
The continuous-approximation experiment follows the same audio fitting protocol
unless stated otherwise, but uses a fixed grid of \(16{,}000\) samples, as
described in the main text. The seen-grid or on-grid evaluation uses the same
coordinate locations used for fitting. The unseen midpoint or inter-sample
evaluation constructs a second grid by placing query points halfway between
adjacent training-grid coordinates. Thus, if the seen grid is
\(\{t_i\}_{i=1}^{N}\), the midpoint grid is
\[
    \tilde{t}_i=\frac{t_i+t_{i+1}}{2},
    \qquad i=1,\ldots,N-1.
\]
This measures whether the learned coordinate function interpolates between
training samples rather than merely matching the discrete training grid. The
same model architecture, optimizer, Fourier-feature construction, and
checkpoint-selection rule are used as in fitting. The appendix includes the full
continuous-approximation sweep over SIREN, FINER, and LSA, including raw-coordinate
and FF16 settings and the corresponding \(\sigma\) values.

\subsection{Images and Derivatives}
\label{app:image_fitting_derivative_ntk}

\paragraph{Kodak data.}
We use the Kodak Lossless True Color Image Suite, using the standard
\texttt{kodim01.png}-\texttt{kodim24.png} images. The dataset contains
24 RGB images, each at spatial resolution \(768\times512\). Images are loaded
as 8-bit RGB PNGs and converted to floating point values in \([0,1]\).
For all coordinate-based experiments, pixel locations are represented by
continuous coordinates
\[
    x=(x_1,x_2)\in[-1,1]^2,
\]
obtained by linearly mapping the horizontal and vertical pixel axes to
\([-1,1]\). Direct image-fitting experiments use the full pixel grid as the
training domain. Optimization is performed with random minibatches of pixels,
while evaluation renders the complete \(768\times512\) grid. Unless explicitly
stated otherwise, direct-fitting train and evaluation grids are therefore the
same grid. No held-out pixels are used for this diagnostic setting.

\paragraph{Image Fitting.}
The image-fitting task learns a coordinate-to-RGB function
\[
    f_\theta:[-1,1]^2\rightarrow \mathbb{R}^3.
\]
For optimization, RGB targets are mapped from $[0,1]^3$ to $[-1,1]^3$. Predictions are mapped back to $[0,1]^3$ before PSNR/SSIM evaluation. All methods in the
official direct-fitting comparison are trained in the no-uplifting regime:
the input to each network is the raw normalized coordinate \(x\), with no
Fourier features, positional encoding, or other input lifting.

The matched-backbone comparison uses width $256$, four nonlinear layers, and a linear RGB output layer. For SIREN we use the official sinusoidal architecture with width \(256\),
\(\omega_0=30\) in the first and hidden layers, and RGB output dimension \(3\).
For FINER we use the official FINER architecture with width \(256\),
\(\omega_0=30\), RGB output dimension \(3\), and first-layer bias scale
\[
    \texttt{first\_bias\_scale}=\sqrt{2}/2 \approx 0.70710678 .
\]
For LSA, each hidden unit applies the learned spectral activation
\[
    \phi(u)
    =
    u+\sum_{k=1}^{K} a_k \sin(2\pi k u),
\]
with \(K=32\) in the main no-Fourier-feature comparison. The coefficients
\(a_k\) are initialized from a zero-mean Gaussian with standard deviation \(10^{-3}\).
We also tested an additive learned DC term on a Kodak subset. It did not improve that diagnostic and is not used in the reported main comparison.

The fixed LSA/FINER/SIREN comparison uses Adam for 5,000 updates, with minibatches of 65,536 pixels, seed 0, and full-grid evaluation every 100 steps. SIREN and FINER use an initial learning rate of $5\times10^{-4}$. LSA ordinary weights and biases also use $5\times10^{-4}$, while its activation coefficients use $10^{-3}$ via the multiplier $\texttt{act\_lr\_mult}=2$. All three models follow the same learning-rate multiplier $0.1^{\min(s/5000,1)}$ at step $s$, applied to both parameter groups for LSA. LSA has 198,915 ordinary MLP weights and biases and $4\times32=128$ activation coefficients, for 199,043 trainable parameters. The LSA calibration candidates and the matched-budget global STAF configuration are specified in \Cref{app:baseline_images}.

The reported checkpoint maximizes evaluated PSNR on the complete fitting grid. This is in-sample signal fitting with no held-out pixels. Image-level holdouts in \Cref{app:baseline_images} concern configuration selection. PSNR is computed on RGB values in $[0,1]$ as
\[
    \mathrm{PSNR}
    =
    -10\log_{10}
    \left(
        \frac{1}{3HW}
        \sum_{p,c}
        \bigl(
            \hat I_{p,c}-I_{p,c}
        \bigr)^2
        +10^{-12}
    \right).
\]
SSIM is computed on RGB images with data range \(1\). In \Cref{fig:kodak_qual}, spatial residuals are mean absolute RGB errors per pixel. Spectral residuals use the channel-averaged signed prediction error $e$, displayed as $\log_{10}(1+|\mathcal F(e)|)$ with a centered two-dimensional FFT. Each residual type uses one 99th-percentile display scale pooled over both images and all three methods.

\paragraph{Gradient-only / Poisson reconstruction.}
For the derivative-supervision experiment, the network is not trained with
direct RGB supervision. Instead, it is trained to match the spatial derivative
field of the target image. We run this experiment on the Kodak images using
the same coordinate domain \(x\in[-1,1]^2\). The main figure uses representative
Kodak examples, and the appendix includes the corresponding residual grids for
the full image set.

Target gradients are computed independently for each RGB channel using
finite differences via \texttt{numpy.gradient}. This gives central finite
differences in the interior and one-sided finite differences at the image
boundary. Let \(I\in[0,1]^{H\times W\times3}\) be the target image. For each
channel \(c\), we compute pixel-index derivatives
\[
    \partial_x I_c,\qquad \partial_y I_c .
\]
Because the INR is differentiated with respect to normalized coordinates in
\([-1,1]^2\), the derivative targets are scaled by the coordinate resolution.
The model output is interpreted in the \([-1,1]\) RGB range during training,
so the derivative targets are
\[
    g^x_{p,c} = (W-1)\,\partial_x I_{p,c},
    \qquad
    g^y_{p,c} = (H-1)\,\partial_y I_{p,c}.
\]
For a sampled coordinate minibatch, gradients of the network output are
computed by automatic differentiation with respect to the input coordinates.
The loss is
\[
    \mathcal{L}_{\mathrm{grad}}
    =
    \left\|
        \partial_x f_\theta(x) - g^x
    \right\|_2^2
    +
    \left\|
        \partial_y f_\theta(x) - g^y
    \right\|_2^2 ,
\]
where the norm is averaged over sampled coordinates and RGB channels. Thus,
RGB derivatives are supervised separately, rather than after conversion to
grayscale.

Derivative-only supervision leaves an additive constant ambiguity. Therefore,
for evaluation and visualization, the predicted image is aligned to the ground
truth by matching its mean RGB value channel-wise:
\[
    \hat I_{\mathrm{aligned}}
    =
    \hat I
    -
    \operatorname{mean}_{p}(\hat I_p)
    +
    \operatorname{mean}_{p}(I_p).
\]
The aligned output is clipped to \([0,1]\), and PSNR is computed after this
mean alignment.

The residual visualizations in the Poisson figures are computed as follows.
The gradient-field residual is
\[
    r_{\nabla}(p)
    =
    \sqrt{
    \frac{1}{3}
    \sum_{c=1}^{3}
    \left[
        \bigl(\partial_x \hat I_{p,c}-\partial_x I_{p,c}\bigr)^2
        +
        \bigl(\partial_y \hat I_{p,c}-\partial_y I_{p,c}\bigr)^2
    \right]
    } .
\]
The image residual is the mean absolute RGB error,
\[
    r_{\mathrm{pix}}(p)
    =
    \frac{1}{3}\sum_{c=1}^{3}
    \left|
        \hat I_{p,c}-I_{p,c}
    \right| .
\]
The spectral residual is computed from the channel-averaged residual
\[
    e(p)=\frac{1}{3}\sum_{c=1}^{3}
    \bigl(\hat I_{p,c}-I_{p,c}\bigr),
\]
using
\[
    r_{\mathrm{fft}}
    =
    \log_{10}
    \left(
        1+
        \left|
            \mathcal{F}(e)
        \right|
    \right),
\]
where \(\mathcal{F}\) is the centered two-dimensional FFT. For display, each
residual type is normalized by a robust \(99\)-th percentile scale computed
over the compared methods within the same figure.

\subsection{Novel View Synthesis}
\label{app:nerf_setup}

We evaluate novel view synthesis on the Blender synthetic dataset~\citep{mildenhall2021nerf}, using the eight scenes
\textit{Chair}, \textit{Drums}, \textit{Ficus}, \textit{Hotdog}, \textit{Lego}, \textit{Materials}, \textit{Mic}, and \textit{Ship}.
We follow the reduced-view protocol used by FINER~\citep{liu2024finer}: from the standard Blender training split, we retain every fourth training image and evaluate on the held-out test views.
Images are rendered and evaluated at downscale factor \(4\), corresponding to \(200\times200\) resolution for the original \(800\times800\) Blender images.
All methods are implemented in the same Instant-NGP-style differentiable volume-rendering codebase, where an MLP predicts density and view-dependent color from 3D position and viewing direction, and pixel colors are obtained by alpha compositing along sampled camera rays.

For all methods we use the same NeRF backbone: four density layers, four color layers, hidden width \(182\), geometry-feature dimension \(182\), and \(512\) samples per ray.
Each optimization step samples \(1024\) rays from the retained training views, with gradient accumulation over four steps, giving an effective ray batch of \(4096\).
Models are trained from scratch for \(150{,}000\) optimization steps using Adam, and reported values use the final checkpoint after this fixed budget.
Only the nonlinear activation family is changed across methods.

SIREN~\citep{sitzmann2020implicit} uses sinusoidal activations with first-layer frequency \(\omega_0=30\) and hidden-layer frequency \(\omega_0=1\).
FINER~\citep{liu2024finer} uses the variable-periodic activation with the same frequency settings and bias scale \(0.577\).
LSA replaces the fixed nonlinearities in the same NeRF MLP with \(K=24\) harmonic components, layer-wise coefficients shared across neurons, and activation-coefficient initialization standard deviation \(5\times10^{-3}\).
We use learning rate \(3\times10^{-4}\) for LSA and \(1\times10^{-4}\) for SIREN and FINER, based on the best stable settings for the shared reduced-view protocol. No scene-specific LSA hyperparameter tuning is used.

We report PSNR, SSIM, and LPIPS on the held-out test views.
PSNR is computed from mean squared RGB error, SSIM from the evaluation code's standard structural-similarity implementation, and LPIPS with the AlexNet backbone.
Qualitative residual maps are computed from saved evaluation tensors as the per-pixel mean absolute RGB error.

\subsection{Neural acoustic fields}
\label{app:naf}

We evaluate LSA in a neural acoustic field setting based on the Neural Acoustic Fields (NAF) formulation~\citep{luo2022learning}. 
Our experiments use the NAF codebase with minimal architectural modification: the baseline model, data loading, spatial feature grid, coordinate embeddings, training protocol, and evaluation pipeline are kept unchanged, except that the hidden activation in the NAF decoder is replaced by LSA. 
We evaluate on the benchmark SoundSpaces/Replica scenes
\texttt{frl\_apartment\_2}, \texttt{frl\_apartment\_4}, \texttt{room\_4}, and \texttt{office\_2}. 
For each scene, the train/test split follows the NAF protocol, where impulse responses are split by emitter-listener spatial queries and held-out queries are used for evaluation. 
Each query consists of a listener position, an emitter position, listener orientation, and binaural channel index. 
Listener and emitter coordinates are represented in the horizontal plane and are combined with sinusoidal embeddings of the time-frequency coordinates. 
As in NAF, the model additionally uses a learned local geometric feature grid queried at the listener and emitter positions. 
The target room impulse responses are represented in the NAF time-frequency output format. The model predicts the normalized spectrogram-domain quantities used by the original NAF implementation, and evaluation converts predictions and ground truth to the corresponding waveform or spectral representations as required by each metric. 
All experiments use the same RIR sample rate, impulse-response length, and normalization files as the NAF benchmark preprocessing.

\paragraph{Architecture.}
The baseline is the shared-grid NAF architecture: listener and emitter positions query a learned spatial feature grid, these features are concatenated with coordinate, orientation, ear, time, and frequency embeddings, and an MLP decoder predicts the NAF output at the queried time-frequency coordinate. 
Let \(u\) denote a scalar hidden pre-activation in this decoder. 
In the LSA variant, we replace the fixed baseline activation by
\begin{equation}
    \phi(u)
    =
    u
    +
    \sum_{k=1}^{K} a_k \sin(2\pi k u),
    \qquad K=8 .
\end{equation}
The LSA coefficients are shared across channels within each layer and are initialized from a zero-mean Gaussian with standard deviation \(10^{-2}\). 
All other NAF modules, including the learned spatial feature grid, coordinate embeddings, orientation and binaural embeddings, output head, and normalization procedure, are unchanged. 
Thus, the comparison isolates the effect of replacing the hidden nonlinear operator while preserving the NAF parameterization and acoustic-field conditioning.

\paragraph{Training.}
We use the original NAF training objective, namely the spectrogram-domain reconstruction loss used by the baseline implementation. 
Models are trained independently per scene. 
Unless otherwise stated, all NAF and LSA-NAF runs use Adam with learning rate \(5\times 10^{-4}\), batch size \(10\), \(256\) sampled time-frequency coordinates per impulse response, hidden width \(512\), and \(50\) epochs. 
For each scene and method, we report the final checkpoint at epoch \(50\). 
The same train/test split, sampled query distribution, preprocessing statistics, and optimization budget are used for the baseline and LSA variants. 
All random seeds are fixed within each paired baseline/LSA comparison whenever the codebase exposes seed control. Otherwise, both methods are run with the same data split and deterministic preprocessing artifacts.

\paragraph{Metrics.}
Metrics are computed on held-out spatial queries and then averaged over spatial positions and scenes. 
Let \(\hat{x}\in\mathbb{R}^{2\times T}\) and \(x\in\mathbb{R}^{2\times T}\) denote the predicted and ground-truth binaural impulse responses after conversion to the waveform domain, with two channels corresponding to the left and right ears.

Waveform MSE is
\begin{equation}
    \mathrm{MSE}(\hat{x},x)
    =
    \frac{1}{2T}\|\hat{x}-x\|_2^2 .
\end{equation}
PSNR is computed from waveform MSE as
\begin{equation}
    \mathrm{PSNR}
    =
    -10\log_{10}\left(\max(\mathrm{MSE},10^{-12})\right),
\end{equation}
assuming unit-normalized waveform scale. 
MAE is the mean absolute waveform error,
\begin{equation}
    \mathrm{MAE}(\hat{x},x)
    =
    \frac{1}{2T}\|\hat{x}-x\|_1 .
\end{equation}
Correlation is the normalized inner product between flattened predicted and ground-truth binaural waveforms,
\begin{equation}
    \mathrm{Corr}(\hat{x},x)
    =
    \frac{\langle \mathrm{vec}(\hat{x}), \mathrm{vec}(x)\rangle}
    {\|\mathrm{vec}(\hat{x})\|_2\|\mathrm{vec}(x)\|_2+\epsilon}.
\end{equation}

For frequency-domain metrics, we compute the real FFT independently for each binaural channel. 
Let \(P=|\mathrm{rFFT}(\hat{x})|\) and \(G=|\mathrm{rFFT}(x)|\). 
Spectral MSE and spectral L1 are
\begin{equation}
    \mathrm{SpectralMSE}
    =
    \frac{1}{|P|}\|P-G\|_2^2,
    \qquad
    \mathrm{SpectralL1}
    =
    \frac{1}{|P|}\|P-G\|_1 .
\end{equation}
Log spectral L1 is
\begin{equation}
    \mathrm{LogSpectralL1}
    =
    \frac{1}{|P|}
    \left\|
    \log(P+\epsilon)-\log(G+\epsilon)
    \right\|_1 .
\end{equation}
For the multi-resolution STFT metric, we compute log-magnitude STFT errors at FFT sizes
\(\{256,512,1024,2048\}\), using a Hann window and hop size \(n_{\mathrm{fft}}/4\) for each resolution, and average the resulting log-L1 errors:
\begin{equation}
    \mathrm{MRSTFT\text{-}LogL1}
    =
    \frac{1}{4}
    \sum_{n\in\{256,512,1024,2048\}}
    \left\|
    \log(|\mathrm{STFT}_{n}(\hat{x})|+\epsilon)
    -
    \log(|\mathrm{STFT}_{n}(x)|+\epsilon)
    \right\|_1 .
\end{equation}
TDOA error measures binaural timing-cue error. 
For each waveform, we estimate the interaural time delay by computing the full cross-correlation between the left and right channels and taking the lag of the maximum response:
\begin{equation}
    \tau(x)
    =
    \arg\max_{\ell}
    \sum_t x_L(t)x_R(t-\ell).
\end{equation}
The TDOA error is
\begin{equation}
    \mathrm{TDOA}(\hat{x},x)
    =
    |\tau(\hat{x})-\tau(x)|,
\end{equation}
reported in samples. 
Lower is better for MSE, MAE, spectral MSE, spectral L1, log spectral L1, MRSTFT log-L1, and TDOA. Higher is better for PSNR and correlation.

We report two aggregate summaries. 
Macro-scene aggregation first averages a metric over spatial positions within each scene and then averages the resulting scene means, giving each scene equal weight. 
Micro-position aggregation pools all evaluated spatial positions across scenes and computes a single average. 
When reporting signed improvements, positive values always indicate that LSA is better: for lower-is-better metrics we use \(\mathrm{baseline}-\mathrm{LSA}\), while for higher-is-better metrics we use \(\mathrm{LSA}-\mathrm{baseline}\).

\paragraph{Spatial improvement maps.}
Figure~\ref{fig:naf_spatial_improvement} visualizes spatially resolved improvements for \texttt{frl\_apartment\_2}. 
Each point corresponds to an evaluated listener/emitter spatial query location projected onto the scene floor plan. 
For lower-is-better metrics, the relative improvement at spatial position \(p\) is
\begin{equation}
    \Delta_{\%}(p)
    =
    100\cdot
    \frac{
    m_{\mathrm{baseline}}(p)-m_{\mathrm{LSA}}(p)
    }{
    \max(|m_{\mathrm{baseline}}(p)|,\epsilon)
    } ,
\end{equation}
so positive values indicate lower error under LSA. 
For higher-is-better metrics, the numerator is reversed. 
Metrics are first averaged over all held-out examples associated with the same spatial position and then visualized on the floor-plan grid. 
For readability, color ranges are clipped symmetrically using a robust percentile of the absolute improvement values for each metric, with a separate colorbar per metric.

\subsection{NTK Energy}
\label{app:ntk_details}
In this section we include implementation details for the experiment from \Cref{fig:motivation_ntk}
Let $f_\theta(x)$ denote the INR and $\{x_i\}_{i=1}^N$ a set of sampled coordinates. 
The empirical neural tangent kernel is defined as
\begin{equation}
\kappa_{ij}
=
\left\langle
\nabla_\theta f_\theta(x_i),
\nabla_\theta f_\theta(x_j)
\right\rangle .
\end{equation}
Let $\kappa = Q\Lambda Q^\top$ be its eigendecomposition, with orthonormal eigenvectors $\{q_i\}$. 
Given a target signal $y\in\mathbb{R}^N$ centered to zero mean, we project it onto the NTK eigenbasis as
\begin{equation}
\alpha_i = \langle q_i,y\rangle .
\end{equation}
We measure the normalized cumulative target energy captured by the top $m$ modes as
\begin{equation}
E(m)
=
\frac{\sum_{i=1}^{m}\alpha_i^2}{\|y\|^2}.
\end{equation}
All methods use the same coordinate subset and target preprocessing. 
All trainable parameters are included in the Jacobian, including LSA activation coefficients.

\subsection{Local Optimization}
\label{app:lsa_theory}

The harmonic coefficients expose a structured family of learning directions. We characterize their contribution to squared-loss descent at a fixed network state, then examine the amplitude parameterization near identity initialization. The results concern the residual on the coordinates used to form the Jacobian.

\subsubsection{Tangent Features and Descent}
Let $X=(x_1,\ldots,x_N)$ be fixed input coordinates and let $f_{w,a}(X)\in\mathbb R^{Nd}$ stack the network outputs, where $d$ is the output dimension. The vector $w$ contains all ordinary weights and biases, and $a$ contains the coefficients $a_k^{(\ell)}$, for $\ell=1,\ldots,L-1$ and $k=1,\ldots,K$. Write $r=f_{w,a}(X)-y$ and $\mathcal L(w,a)=\tfrac12\|r\|_2^2$. A mean-squared-error normalization can be absorbed into the group learning rates.

For one input, let $B_\ell(x)=\partial f(x)/\partial z^{(\ell)}(x)$ be the downstream Jacobian from layer $\ell$ to the output. Since $a_k^{(\ell)}$ is shared across channels, its output tangent feature is
\begin{equation}
\psi_{\ell k}(x)=B_\ell(x)\sin\!\left(2\pi k u^{(\ell)}(x)\right),
\label{eq:propagated_harmonic}
\end{equation}
where the sine acts elementwise and the matrix multiplication sums the channel contributions. Stacking these vectors over $X$ gives the corresponding column of $\Psi=\partial f(X)/\partial a$. Let $J_w=\partial f(X)/\partial w$.

Under full-batch gradient flow with fixed positive rates $\eta_w$ and $\eta_a$,
\begin{equation}
\dot w=-\eta_w J_w^\top r,\qquad
\dot a=-\eta_a\Psi^\top r,\qquad
\dot r=-\left(\eta_wJ_wJ_w^\top+\eta_a\Psi\Psi^\top\right)r.
\label{eq:group_flow}
\end{equation}
The chain rule yields
\begin{equation}
-\dot{\mathcal L}
=\eta_w\|J_w^\top r\|_2^2+\eta_a\|\Psi^\top r\|_2^2
=\eta_w\|J_w^\top r\|_2^2+\eta_a\sum_{\ell,k}\langle r,\psi_{\ell k}\rangle^2.
\label{eq:appendix_descent}
\end{equation}
This identity is exact at finite width, including when the tangent features change during training. Each coefficient follows the correlation of the residual with its downstream-propagated harmonic feature. At a common network state, the additional instantaneous descent contribution from updating $a$ is exactly $\eta_a\|\Psi^\top r\|_2^2$. It is positive when the residual has a nonzero projection onto $\operatorname{col}(\Psi)$.

In the fixed-NTK approximation, evaluate both Jacobians at the shared initialization $(w_0,a_0)$ and keep them fixed. The model with frozen coefficients trains $w$ while holding $a=a_0$. Its effective kernel and that of the jointly trained model satisfy
\begin{equation}
\kappa_{\mathrm{frozen},0}=\eta_wJ_{w,0}J_{w,0}^\top,\qquad
\kappa_{\mathrm{LSA},0}=\kappa_{\mathrm{frozen},0}+\eta_a\Psi_0\Psi_0^\top.
\end{equation}
The update is positive semidefinite and has rank at most $(L-1)K$, the number of coefficients. For either fixed kernel $\kappa_0$, the residual evolves as $r(t)=\exp(-\kappa_0t)r_0$. At initialization, the difference in loss-decay rates is $\eta_a\|\Psi_0^\top r_0\|_2^2$. These comparisons hold at a shared state. Subsequent nonlinear trajectories develop their own Jacobians and residuals, so the applicable guarantee is instantaneous descent and its fixed-kernel local approximation. Generalization to unobserved coordinates is evaluated empirically in the audio and neural-field experiments.

The harmonic choice determines the alignment of the added directions. In a simplified local model with orthonormal propagated harmonic features $q_1,\ldots,q_K$, let $r=\sum_k c_kq_k+r_\perp$, where $r_\perp$ is orthogonal to their span. For an equal-weight adapter kernel $\kappa_a=\lambda\sum_kq_kq_k^\top$, its target-weighted contribution is $r^\top\kappa_ar=\lambda\sum_kc_k^2$. This describes how a fixed kernel-trace budget can concentrate on harmonic residual energy. For the actual propagated features, the exact quantity is $\|\Psi^\top r\|_2^2$, which also reflects their scale and correlation. The resulting inductive bias is target-dependent.

\subsubsection{Amplitude Conditioning}
Consider one residual sinusoidal term $g(u)=u+A\sin(\omega u+\varphi)$. Its parameter derivatives are
\begin{equation}
\frac{\partial g}{\partial A}=\sin(\omega u+\varphi),\qquad
\frac{\partial g}{\partial\omega}=Au\cos(\omega u+\varphi),\qquad
\frac{\partial g}{\partial\varphi}=A\cos(\omega u+\varphi).
\label{eq:amplitude_derivatives}
\end{equation}
Assume a fixed finite network and evaluation set, bounded pre-activations and downstream Jacobians as $A\to0$, and fixed frequencies and phases. The chain rule then gives frequency and phase output-Jacobian blocks of order $O(A)$, and their Gram matrices are $O(A^2)$. The amplitude block is $O(1)$, and has a nonzero limit whenever the propagated sine features are nondegenerate. The same argument holds for a finite bank with amplitudes $A_k=\epsilon c_k$ and bounded $c_k$ as $\epsilon\to0$.

For LSA, $\omega_k=2\pi k$ and $\varphi_k=0$. At exact-zero coefficient initialization,
\begin{equation}
\phi(u,0)=u,\qquad
\frac{\partial\phi}{\partial u}(u,0)=1,\qquad
\frac{\partial\phi}{\partial a_k}(u,0)=\sin(2\pi ku).
\end{equation}
The activation and its input derivative preserve the identity, while coefficient learning can access harmonic output directions. This motivates amplitude learning under an identity residual parameterization. STAF also learns amplitudes, and its full parameterization and initialization are evaluated separately in \Cref{app:additional_baselines}.

\subsubsection{Controlled Validation}
We validated the finite-width identity using six Kodak images and three network initializations per image. The model has four coefficient-bearing layers with $K=32$, giving a coefficient-kernel rank bound of 128. The learned and frozen variants begin from the same state. For a common outer gradient-descent step size $\alpha$, define the one-step loss difference normalized by its first-order prediction as
\begin{equation}
R(\alpha)=\frac{\mathcal L_{\mathrm{frozen}}(\alpha)-\mathcal L_{\mathrm{learned}}(\alpha)}{\alpha\eta_a\|\Psi_0^\top r_0\|_2^2}.
\label{eq:finite_step_ratio}
\end{equation}
Here $\mathcal L_{\mathrm{learned}}(\alpha)$ follows one joint update from the initial state, and $\mathcal L_{\mathrm{frozen}}(\alpha)$ follows the corresponding ordinary-weight update only. Differentiability gives $R(\alpha)\to1$ as $\alpha\to0$ whenever the denominator's coefficient is positive. With $\alpha_0=0.1/\lambda_{\max}(\kappa_{\mathrm{LSA},0})$ and equal group rates, the measured means are shown in \Cref{tab:finite_step}. The smallest tested step has a positive difference in all 18 image-seed pairs. Exact-zero initialization reaches a mean full-image PSNR within 0.044 dB of small-random initialization on the six-image comparison.

\begin{table}[htbp]
\centering
\caption{One-step validation on six Kodak images and three initializations per image. The normalization in \Cref{eq:finite_step_ratio} tests the local descent prediction.}
\label{tab:finite_step}
\begin{tabular}{lccc}
\toprule
$\alpha/\alpha_0$ & 0.10 & 0.03 & 0.01 \\
\midrule
Mean $R(\alpha)$ & 0.806 & 0.942 & 0.981 \\
\bottomrule
\end{tabular}
\end{table}

For amplitude conditioning, we evaluated final-output Jacobians of a scalar-input, scalar-output network with width 32, three residual sinusoidal layers, and $K=8$. We used 512 uniformly spaced coordinates on $[0,1]$, three ordinary-weight seeds, and float64 Jacobian calculations. Ordinary weights were fixed across the amplitude sweep. With $A_k=\epsilon$, $\omega_k=2\pi k$, and zero phases, log-log fits over $\epsilon\in[10^{-6},10^{-2}]$ gave squared frequency-Jacobian exponents from 2.18 to 2.27 and squared phase-Jacobian exponents near 2.00. Amplitude exponents remained near zero, and the output perturbation scaled approximately linearly with $\epsilon$. At $\epsilon=0$, the amplitude block remained nonzero for every tested seed.

We also compared amplitude adapters with the same coefficient count and matched kernel trace in this network. All adapters start with zero amplitudes and identical ordinary parameters. The harmonic adapter uses $\sin(2\pi ku)$. The random-frequency adapter uses $\sin(2\pi\nu_ku+\varphi_k)$ with $\nu_k\sim\mathcal U[0.5,8.5]$ and $\varphi_k\sim\mathcal U[0,2\pi]$. The jittered adapter uses $\sin(2\pi(k+\delta_k)u)$ with $\delta_k\sim\mathcal U[-0.4,0.4]$. Five dictionary seeds are evaluated for each stochastic family. For each ordinary-weight seed, coefficient kernels are rescaled to the harmonic kernel's trace. Let $\widetilde\kappa_a$ denote this trace-matched kernel and define $D_a=r_0^\top\widetilde\kappa_ar_0$.

On the zero-mean, unit-RMS edge target $y(x)\propto\tanh(12\sin(2\pi\,3x))$, the harmonic adapter attains $D_a=0.0611$. The five random dictionaries span $0.00020$ to $0.00481$, and the mean jittered result is $0.0616$. The same ordering holds for the tested spectral-stack target. This comparison associates the observed gain with coordinated harmonic structure under a controlled kernel scale. The local benefit remains determined by the target-weighted alignment of the propagated features.

\section{Additional Results}
\label{app:addres}
In this section we bring additional results for the experiments discussed in \Cref{sec:results}.
\subsection{Audio Data}
\subsubsection{Audio Signal Fitting}
\label{app:asf}
\begin{figure}[t]
\centering
\includegraphics[width=\linewidth]{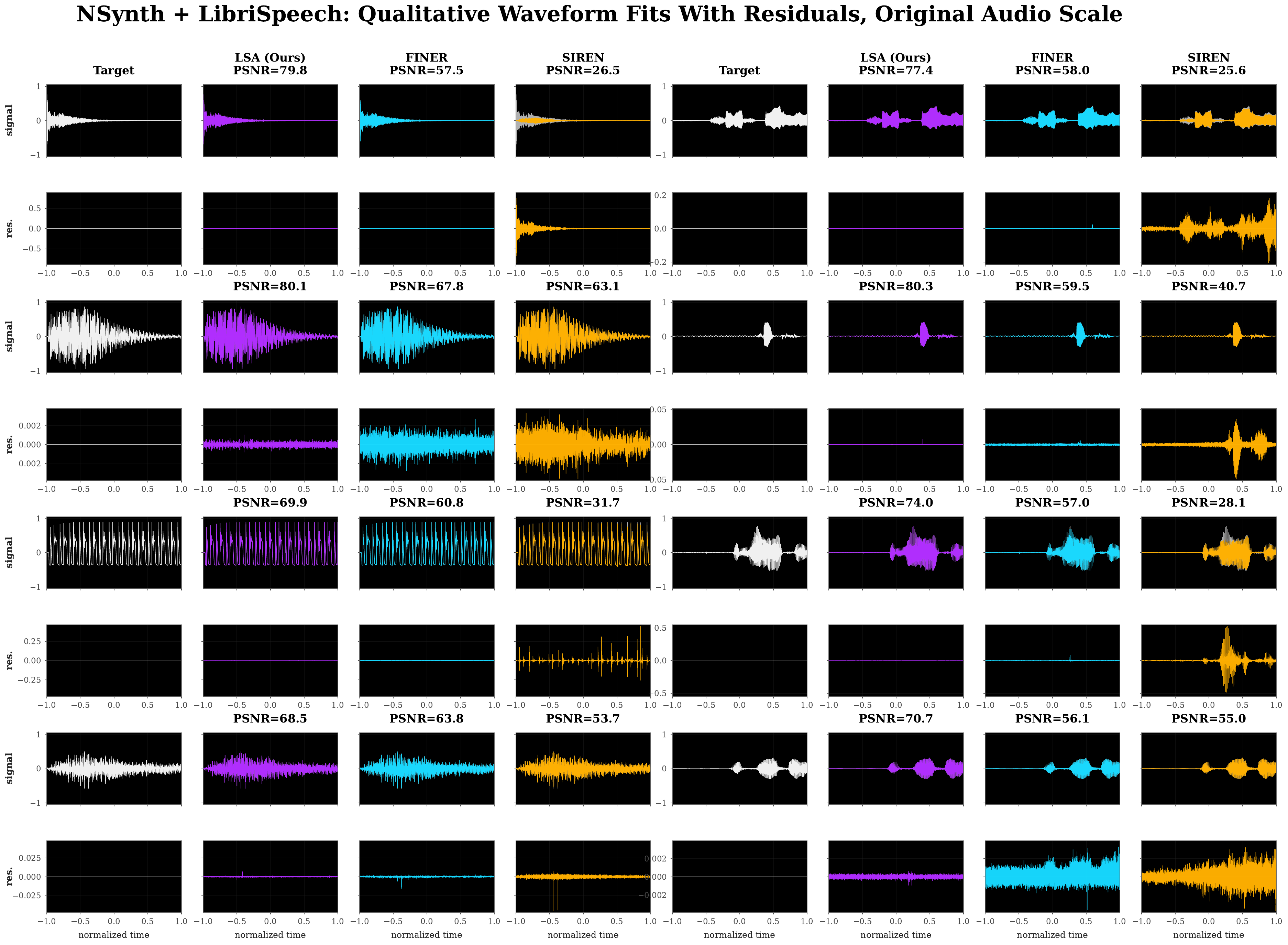}
\caption{
Qualitative waveform fits on LibriSpeech and NSynth. Each column compares the target with a reconstruction and its residual on the original audio scale. Learned spectral activations closely match transient and harmonic waveform structure, leaving substantially smaller residuals than FINER and SIREN.
}
\label{fig:audio_qual}
\end{figure}

\Cref{fig:audio_qual} shows representative waveform fits and residuals. On LibriSpeech, learned spectral activations nearly overlap the target waveforms in the shown windows, whereas FINER leaves visible errors around transient regions and SIREN often misses amplitude and fine waveform structure. On NSynth, learned spectral activations preserve both periodic structure and decaying envelopes, while FINER captures the coarse oscillatory pattern but leaves larger residuals, and SIREN shows the largest mismatch.

These qualitative differences match the operator-level motivation in \Cref{fig:operator_motivation}. Speech transients and musical timbre are not merely high-frequency signals. They require coordinated harmonic structure whose amplitudes and envelopes vary over time. Frequency warping alone can expose or shift oscillatory components, but it does not directly parameterize their composition. Learned spectral activations, through \Cref{eq:method_activation}, provide this missing degree of freedom by learning the harmonic mixture of the hidden nonlinear response.

\subsubsection{Continuous Approximation}
For the main audio fitting experiments, one-second clips are represented at
48 kHz, yielding \(48{,}000\) samples per clip. The continuous-approximation
experiment is separate: it uses a fixed grid of \(16{,}000\) training samples
and evaluates both on-grid and at unseen midpoints between adjacent grid points.
\begin{figure}[t]
\centering
\includegraphics[width=\linewidth]{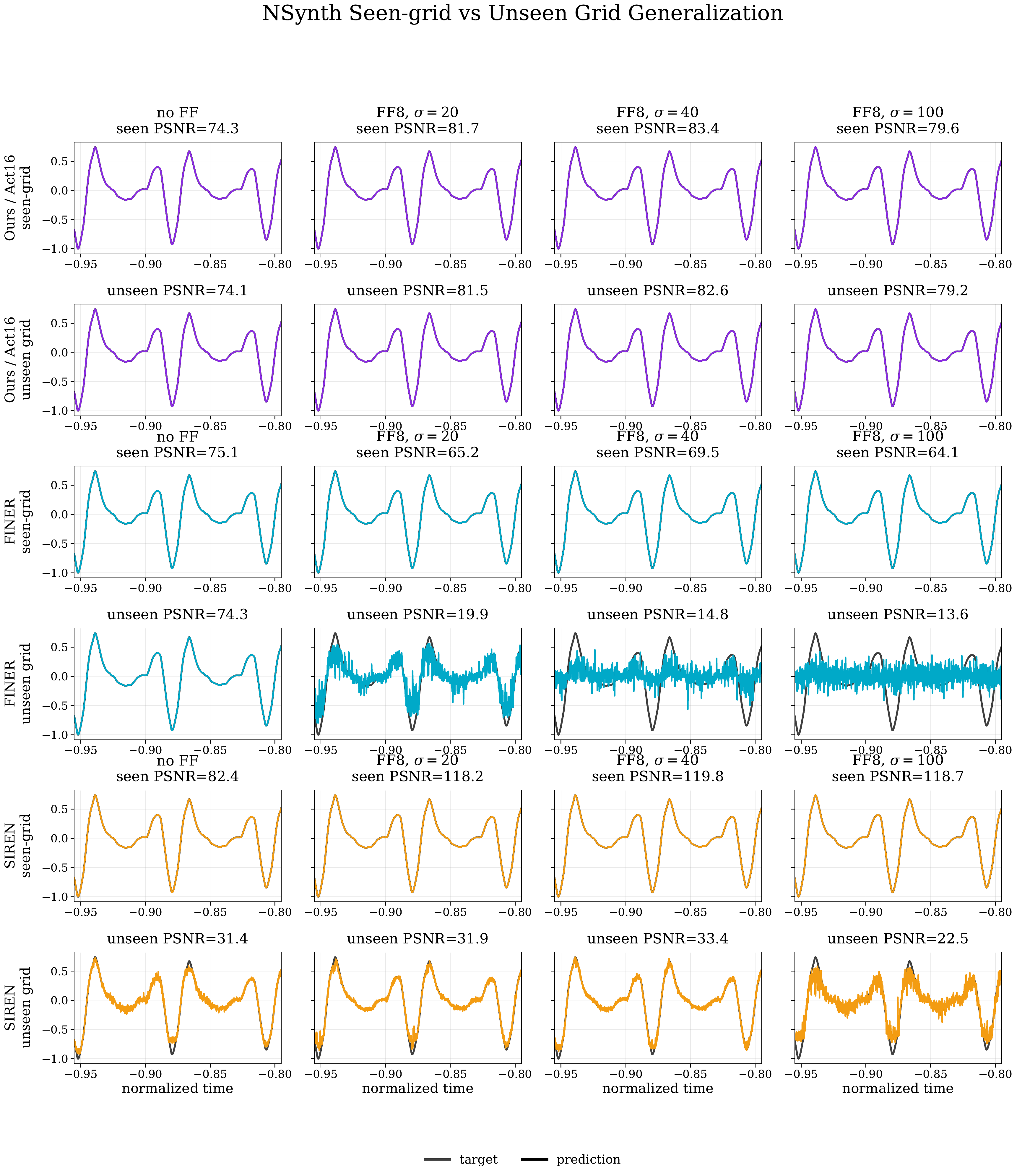}
\caption{Full NSynth on-grid vs. off-grid diagnostic across bandwidth settings.}
\label{fig:nsynth_seen_unseen_full}
\end{figure}

\subsection{Image Fitting}
\label{app:image_res}
\begin{figure}[t]
    \centering
    \includegraphics[width=\linewidth]{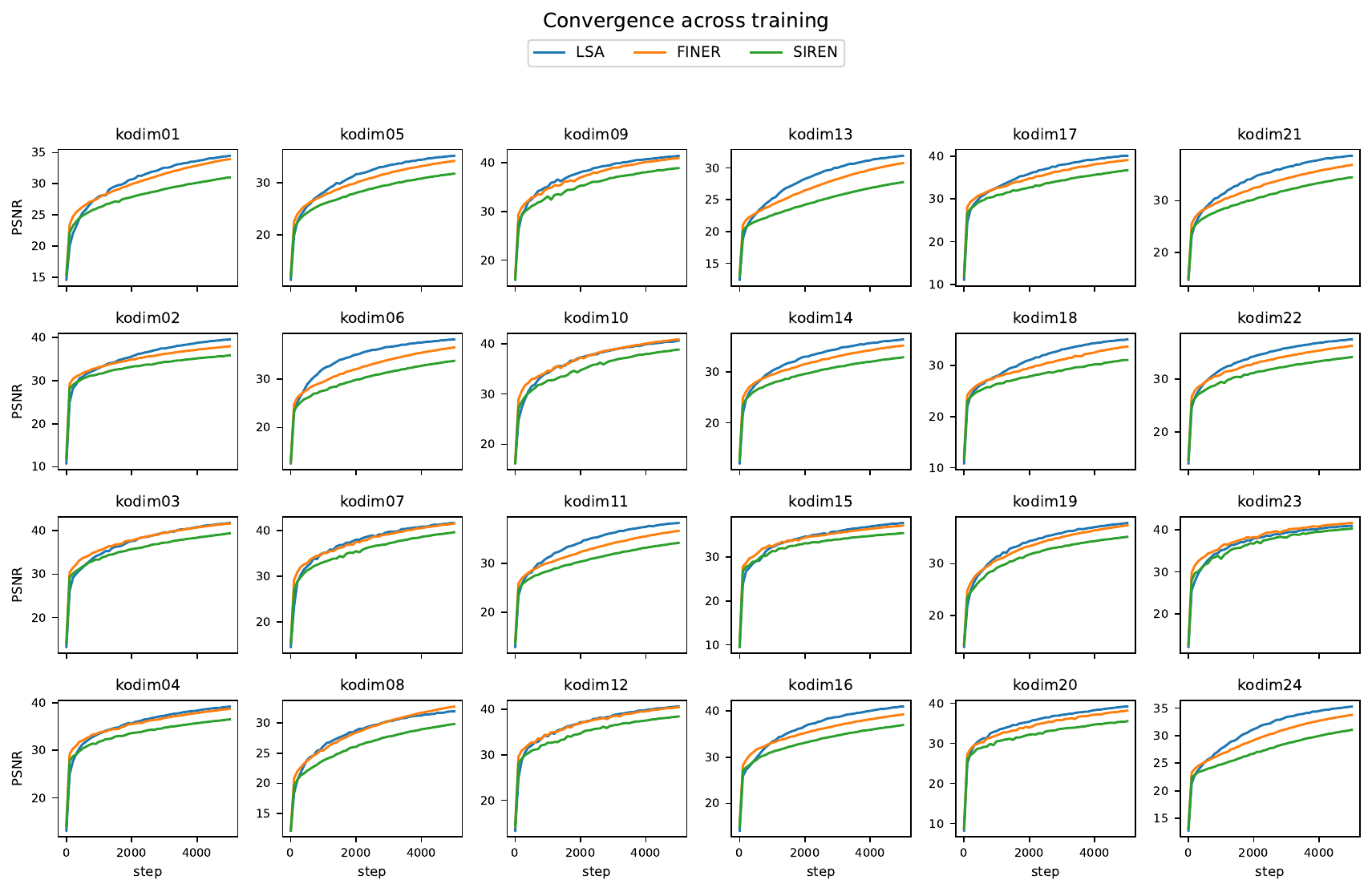}
    \caption{
    Convergence curves for all 24 Kodak images under fixed LSA, FINER, and SIREN configurations and matched 5,000-update budgets. LSA separates from SIREN early, while its advantage over FINER emerges progressively. Early-stage behavior varies across images.
    }
    \label{fig:kodak_convergence_full}
\end{figure}

\paragraph{Additional convergence diagnostics.}
\Cref{fig:kodak_convergence_full} shows PSNR trajectories for all Kodak images under the fixed configurations and 5,000-update budget in \Cref{sec:results_images}. At steps 500, 1,000, 2,000, 3,000, 4,000, and 5,000, LSA leads FINER on 5, 16, 21, 22, 21, and 21 of 24 images, respectively. It leads SIREN on 23/24 images at step 500 and all 24 at the remaining listed steps. Thus LSA separates from SIREN early, while its advantage over FINER emerges progressively. Early-stage behavior varies across images.

\subsection{Neural Fields}
\label{app:nf}

\begin{figure}[t]
    \centering
    \includegraphics[width=\linewidth]{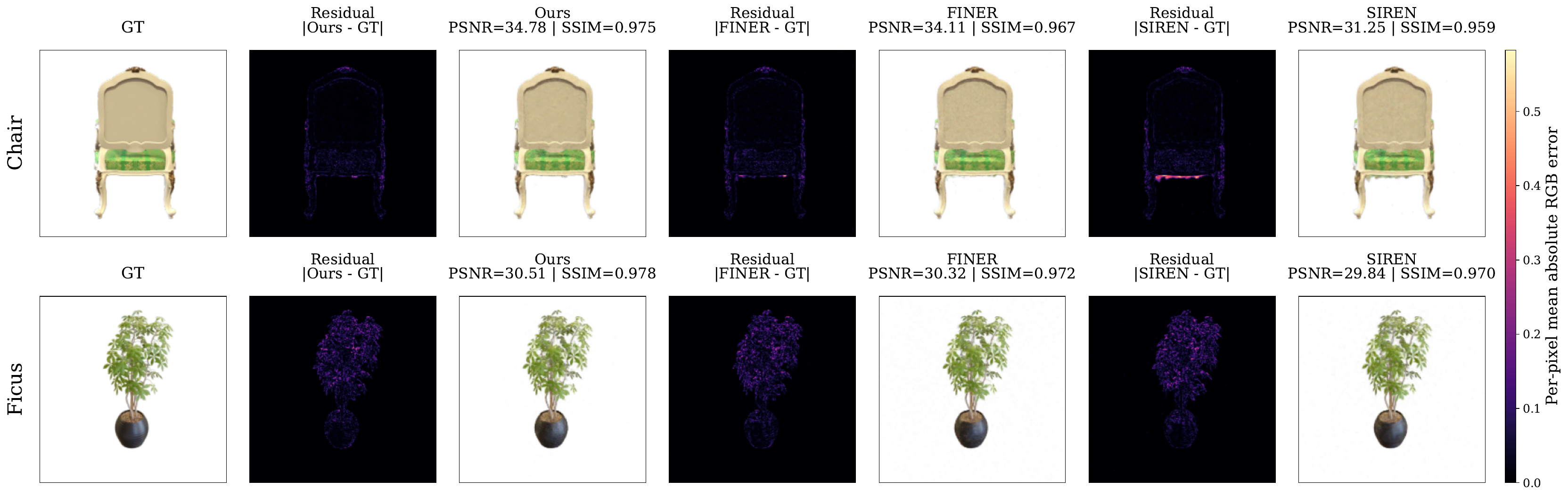}
    \caption{Qualitative novel view synthesis results. We show target views, reconstructions, and absolute RGB residuals for representative scenes. Learned spectral activations reduce structured residuals around fine object boundaries and high-detail regions compared to SIREN and FINER.}
    \label{fig:nerf_qualitative}
\end{figure}

\paragraph{NeRFs.} 
The qualitative comparison in \Cref{fig:nerf_qualitative} supports the interpretation from \Cref{sec:results_nerf}. On Chair and Ficus, all methods recover the coarse object appearance, but the residual maps show that SIREN and FINER retain more structured errors around thin geometry, object boundaries, and textured regions. LSA reduces these residuals while preserving comparable global appearance. This aligns with the mechanism in \Cref{fig:operator_motivation}: neural rendering does not only require access to high frequencies, but efficient composition of localized appearance and geometric detail within the internal representation.

\subsection{Neural-Acoustic Fields}
We evaluate on benchmark scenes used in NAF~\cite{luo2022learning}, including \texttt{frl\_apartment\_2}, \texttt{frl\_apartment\_4}, \texttt{room\_4}, and \texttt{office\_2}. 
\Cref{tab:naf_lsa} reports aggregate results using two complementary summaries: macro-scene averaging, which weights each scene equally, and micro-position averaging, which aggregates over all evaluated spatial positions.

LSA improves every reported metric under both aggregation schemes. The largest gains appear in metrics that are sensitive to the time-frequency structure of the impulse response: log spectral L1 improves by $44.90\%$ under macro-scene aggregation and $50.84\%$ under micro-position aggregation, while MRSTFT log-L1 improves by $45.97\%$ and $51.54\%$, respectively. The improvement in TDOA is even larger, reaching $67.55\%$ and $72.41\%$, indicating better recovery of spatial timing cues. Waveform-domain metrics also improve consistently, with MSE reduced by approximately $9$-$10\%$ and correlation increasing by roughly $40$-$43\%$.
\begin{table}[t]
\centering
\caption{Neural acoustic field results. We replace the activation in the NAF model with LSA ($K=8$) while keeping the architecture and training protocol fixed. Improvements are signed so that positive values indicate better performance.}
\label{tab:naf_lsa}
\resizebox{\linewidth}{!}{
\begin{tabular}{lcccccc}
\toprule
& \multicolumn{3}{c}{Macro-scene aggregation} & \multicolumn{3}{c}{Micro-position aggregation} \\
\cmidrule(lr){2-4}\cmidrule(lr){5-7}
Metric & Baseline & LSA & Improvement & Baseline & LSA & Improvement \\
\midrule
PSNR from MSE $\uparrow$ & 11.417 & 11.846 & +0.429 / +3.76\% & 11.550 & 11.993 & +0.443 / +3.84\% \\
MSE $\downarrow$ & 0.07238 & 0.06572 & +0.0067 / +9.20\% & 0.07 & 0.063 & +0.0068 / +9.71\% \\
Log Spectral L1 $\downarrow$ & 4.053 & 2.233 & +1.820 / +44.90\% & 3.981 & 1.957 & +2.024 / +50.84\% \\
MRSTFT Log-L1 $\downarrow$ & 4.008 & 2.165 & +1.842 / +45.97\% & 3.925 & 1.902 & +2.023 / +51.54\% \\
TDOA $\downarrow$ & 122.406 & 39.727 & +82.679 / +67.55\% & 114.675 & 31.634 & +83.041 / +72.41\% \\
Spectral L1 $\downarrow$ & 25.558 & 22.061 & +3.497 / +13.68\% & 25.252 & 21.173 & +4.079 / +16.15\% \\
Spectral MSE $\downarrow$ & 975.873 & 770.363 & +205.510 / +21.06\% & 954.076 & 719.924 & +234.152 / +24.54\% \\
MAE $\downarrow$ & 0.2147 & 0.2029 & +0.0118 / +5.49\% & 0.2141 & 0.2008 & +0.0134 / +6.24\% \\
Corr $\uparrow$ & 0.2740 & 0.3823 & +0.1083 / +39.52\% & 0.2815 & 0.4039 & +0.1224 / +43.47\% \\
\bottomrule
\end{tabular}
}
\end{table}

\paragraph{Additional Spatial Improvement Maps.}
We bring results similar to \Cref{fig:naf_spatial_improvement}, conclusions from \Cref{sec:results_nerf} stably reproduce across different scenes. 
\begin{figure}[t]
    \centering
    \includegraphics[width=\linewidth]{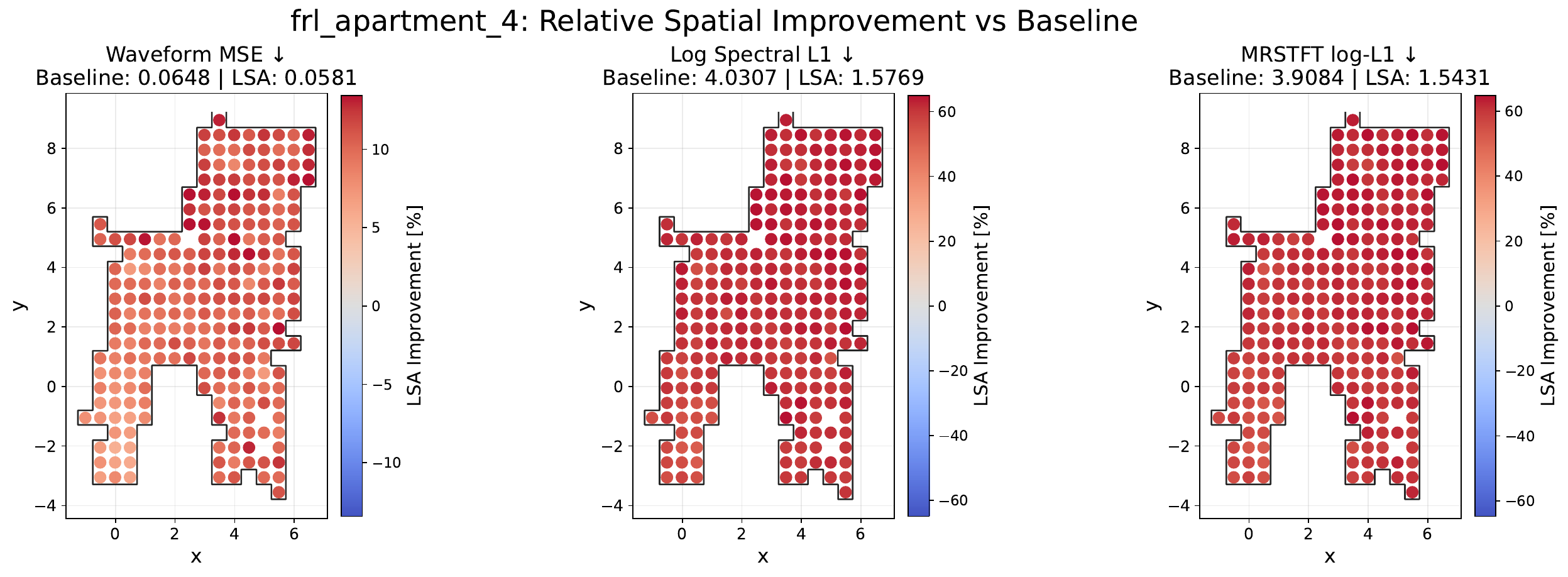}
    \caption{Spatial improvement of LSA over the NAF baseline. frl apartment scene 4}
    \label{fig:naf_spatial_improvement_frl_4}
\end{figure}
\begin{figure}[t]
    \centering
    \includegraphics[width=\linewidth]{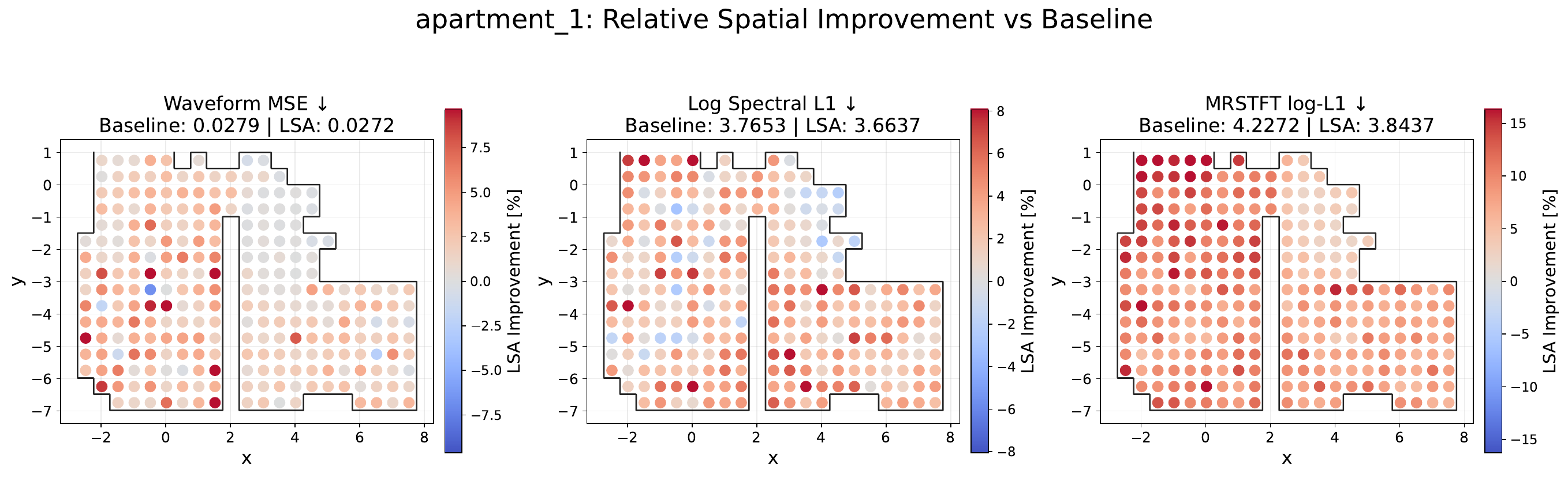}
    \caption{Spatial improvement of LSA over the NAF baseline. Apartment Scene 1}
    \label{fig:naf_spatial_improvement_apartment_1}
\end{figure}\begin{figure}[t]
    \centering
    \includegraphics[width=\linewidth]{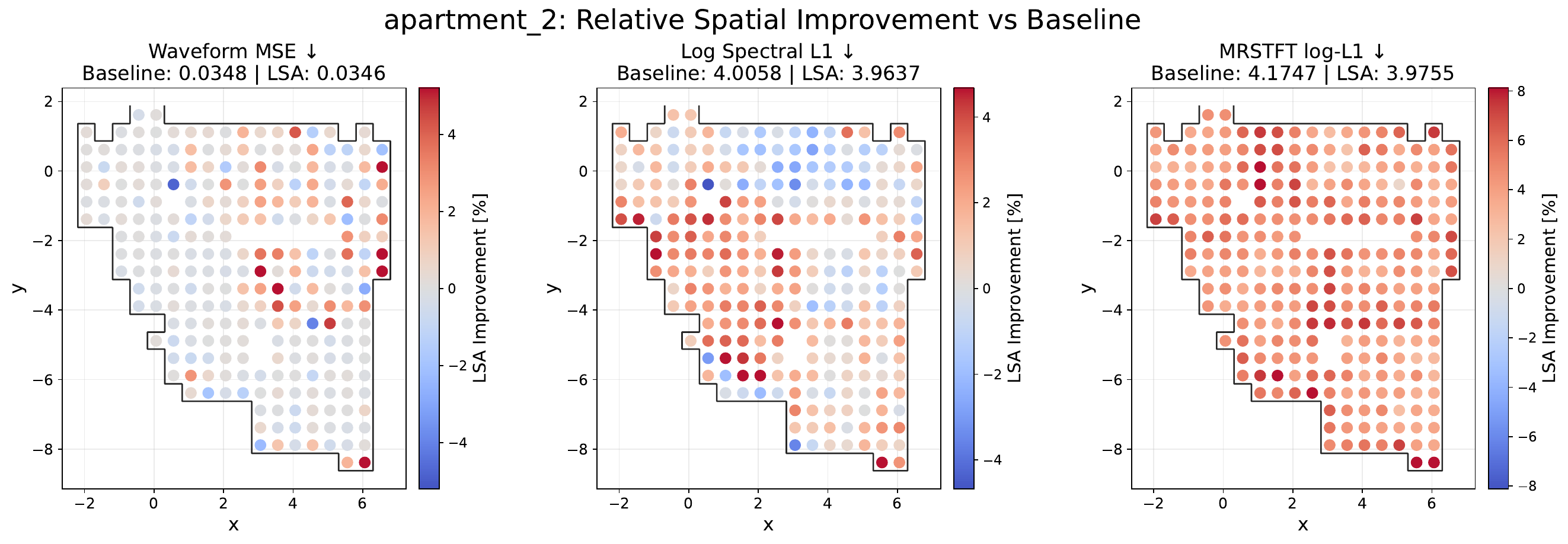}
    \caption{Spatial improvement of LSA over the NAF baseline. Apartment Scene 2.}
    \label{fig:naf_spatial_improvement_apartment_2}
\end{figure}
\begin{figure}[t]
    \centering
    \includegraphics[width=\linewidth]{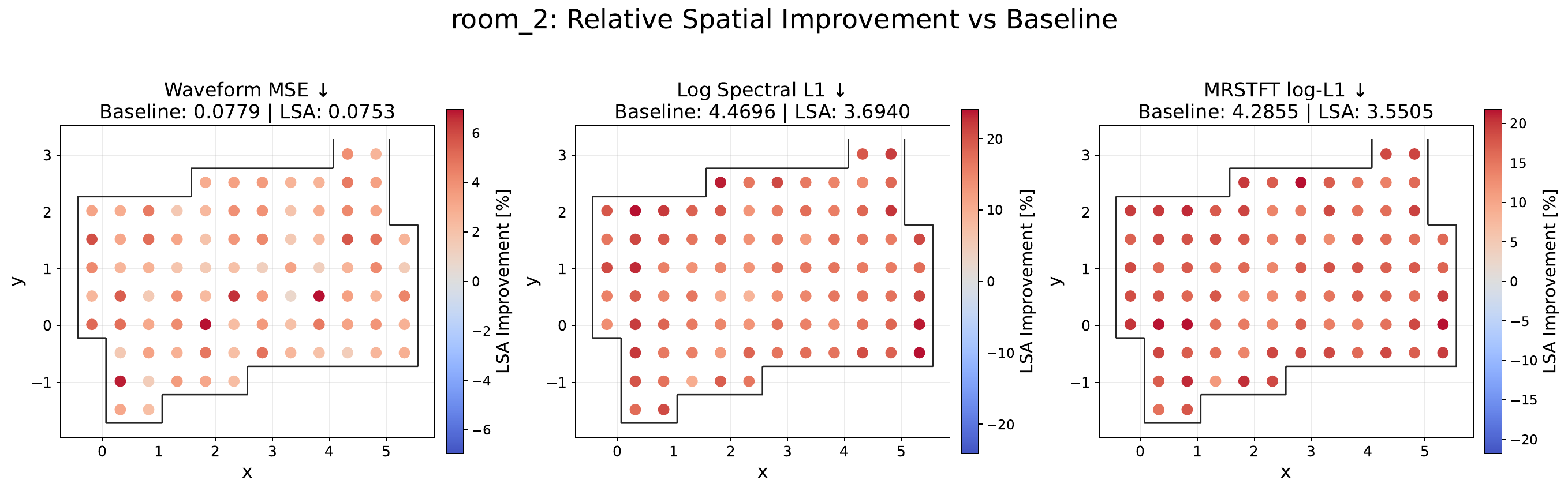}
    \caption{Spatial improvement of LSA over the NAF baseline.Room Scene.}
    \label{fig:naf_spatial_improvemen_roomt}
\end{figure}

\begin{figure}[t]
    \centering
    \includegraphics[width=\linewidth]{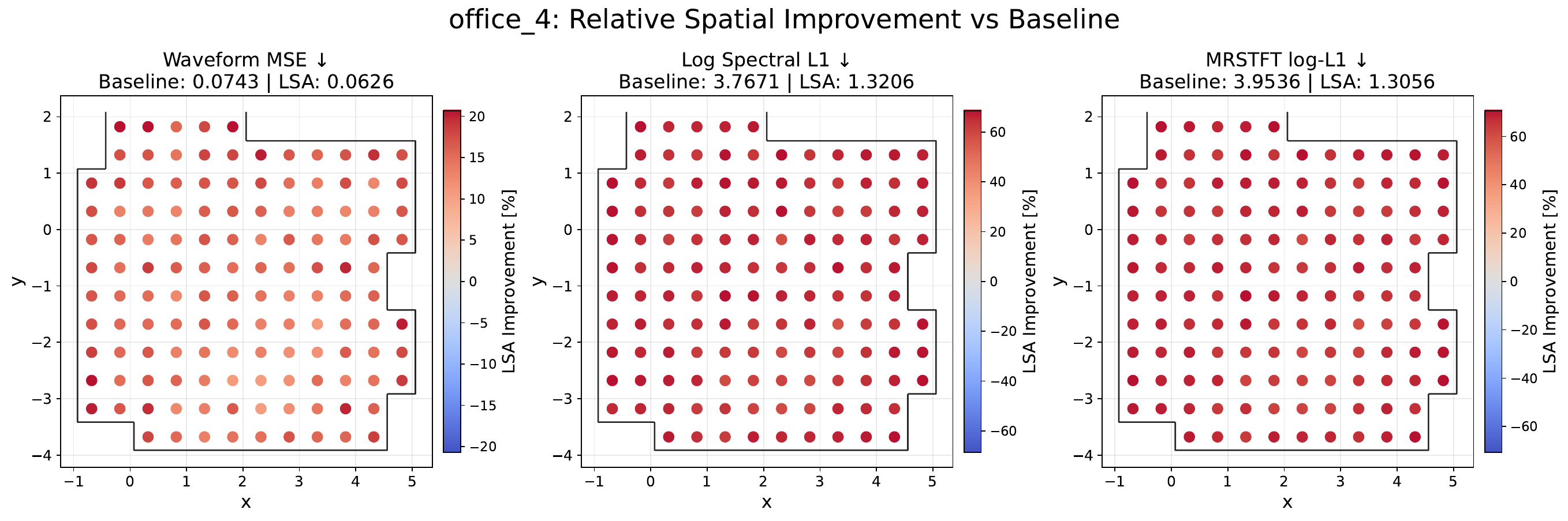}
    \caption{Spatial improvement of LSA over the NAF baseline. Office Scene}
    \label{fig:naf_spatial_improvement_office}
\end{figure}

\subsection{Additional Use-cases}
\label{app:pe}

While this paper focuses on the application of learned spectral activations for INRs, in this section we display initial evidence for the potential of our method beyond this scope. We evaluate positional encodings (PEs) in a sparse-to-dense reconstruction task, where a Transformer predicts the full image from a small set of observed pixels and their coordinates. Unlike dense INR fitting, this setting requires reasoning about spatial relationships between observed and query locations, making the choice of PE critical. All models share the same architecture and training setup, only the PE differs - we compare between not using any form of PE, the seminal sinusoidal PE from \cite{vaswani2017attention}, plain coordinate-space learning of PE, and learned PE based on our proposed spectral formulation (see Appendix). To ablate the amount of signal carried by the coordinate identity rather than its RGB values, we also compare to a variant of our learned PE that does not receive the raw pixel data as input, but only coordinate location. For all regimes we report PSNR high-gradient PSNR (\textit{Edge}), computed only on the top 20\% of pixels by ground-truth gradient magnitude.

\begin{figure}[t]
\centering
\includegraphics[width=\linewidth]{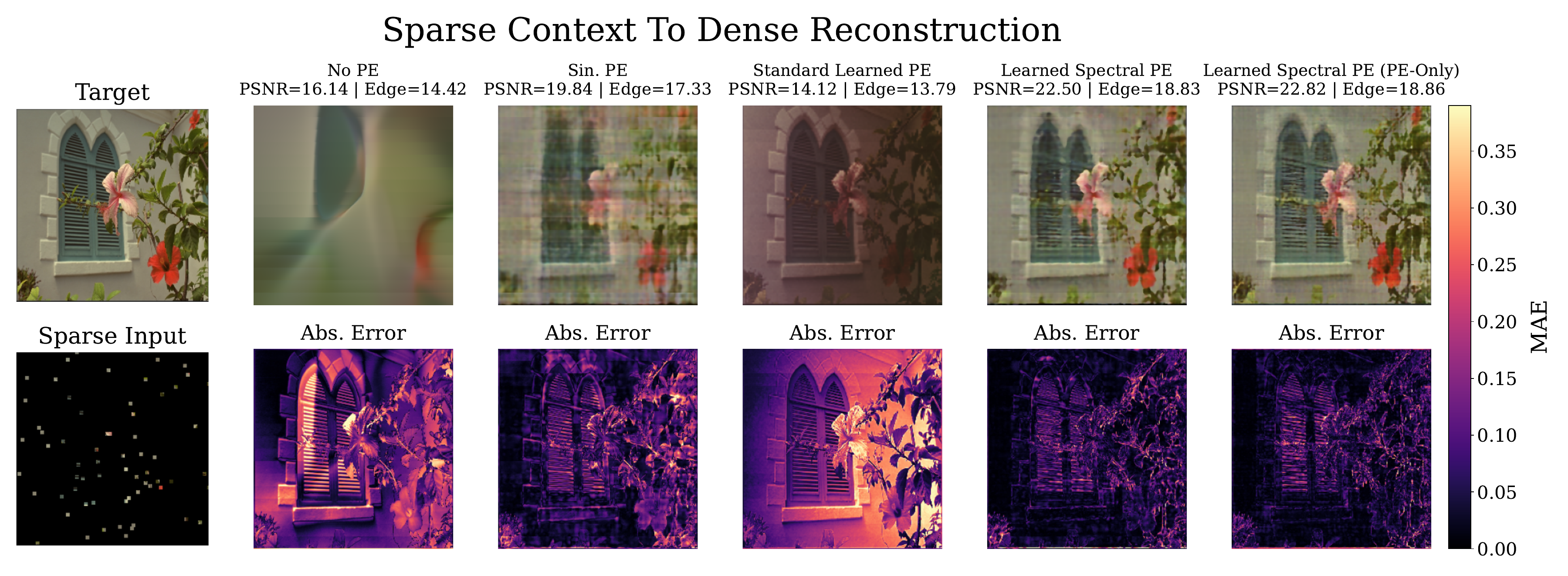}
\caption{Sparse context $\rightarrow$ dense reconstruction under different positional encodings. Top: reconstructions. Bottom: absolute error maps.}
\label{fig:pe_sparse_recon}
\end{figure}

Figure~\ref{fig:pe_sparse_recon} shows that the task is highly PE-sensitive. Removing structure (\emph{No PE}) yields oversmoothed reconstructions, while learned table embeddings perform poorly, indicating that unstructured learned PEs are insufficient. Fixed sinusoidal PE improves results but exhibits structured artifacts and elevated residuals near edges and fine details.

Learned spectral PE consistently produces sharper reconstructions and lower errors, particularly in high-frequency regions. Notably, the \emph{PE-only} variant-where spatial information is provided exclusively through the learned spectral encoding-performs best, showing that the learned representation itself forms an effective coordinate system.

From the harmonic dictionary perspective (\Cref{sec:method}), fixed PEs impose a predefined spectral basis, while learned spectral PE adapts this basis to the task. The improvement observed here indicates that learning the spectral structure of positional encodings yields a more suitable representation of spatial relationships, rather than merely adding features.

\clearpage
\section{Additional Comparisons}
\label{app:additional_baselines}

These experiments report matched-backbone activation comparisons and broader INR comparisons under their respective protocols. Each comparison specifies its tuning and evaluation protocol. STAF~\citep{morsali2026staf} and SL$^2$A~\citep{rezaeian2025sl2a} use their released implementations. WIRE~\citep{saragadam2023wire} and MFN GaborNet~\citep{fathony2020multiplicative} provide wavelet and multiplicative-filter comparisons. The fractional Jacobi experiments use the single-degree sequential fJNB construction of~\citet{aghaei2024fkan}. We use the label fJNB throughout these tables to distinguish that construction from the Fourier KAN of~\citet{mehrabian2025fourier}, discussed in \Cref{sec:related}.

\subsection{Image Fitting}
\label{app:baseline_images}

We evaluate full-resolution Kodak images using raw normalized coordinates and RGB outputs. The controlled MLP comparisons use width 256, four nonlinear layers, minibatches of 65,536 pixels, seed 0, and the image metrics in \Cref{app:image_fitting_derivative_ntk}. We report two selection protocols: per-image tuning and a fixed configuration selected on six calibration images, then evaluated on the other 18 images. The calibration images are kodim02, kodim05, kodim07, kodim10, kodim13, and kodim23. Under the latter protocol, every image is fitted separately. The holdout concerns hyperparameter selection across images.

LSA calibration tests the $(\mathrm{learning\ rate},\mathrm{coefficient\ multiplier})$ pairs $(5\times10^{-4},2)$, $(10^{-3},2)$, and $(10^{-3},1)$. All use $K=32$, 5,000 updates, and exponential decay to 0.1 of the initial rate. The selected global configuration, $(5\times10^{-4},2)$, is frozen for \Cref{fig:kodak_bar,fig:kodak_qual,fig:kodak_convergence_full}. Per-image LSA tuning selects separately from these same three candidates.

For the paired PSNR intervals in \Cref{tab:staf_per_image_summary,tab:sl2a_kodak,tab:kodak_holdout_extra}, we resample paired image-level differences 20,000 times and report the 2.5th and 97.5th percentiles of the bootstrap means. Means and paired gains are rounded independently.

\paragraph{STAF.}
We use the released skip-connected STAF implementation with first-layer $\omega_0=60$, hidden-layer $\omega_0=30$, and its other default settings. The per-image sweep comprises 48 configurations with $\tau\in\{3,5,8,16\}$ sinusoidal terms, learning rates $\{2.5\times10^{-4},5\times10^{-4},10^{-3},2\times10^{-3}\}$, and constant or decayed schedules with final multipliers 0.1 and 0.05. An additional eight configurations use $\tau=3$, learning rates $\{3\times10^{-3},5\times10^{-3},10^{-2},2\times10^{-2}\}$, and the two decayed schedules. Thus STAF receives 56 candidates per image, with 5,000 updates per fit. LSA uses three candidates per image. The boundary extension improves STAF on 22 of 24 images, by 0.545 dB on average. Selection uses each image's best evaluated fitting PSNR. \Cref{tab:staf_per_image_summary} summarizes the selected results, and \Cref{tab:staf_per_image} gives the paired image values.

\begin{table}[htbp]
\centering
\caption{Per-image Kodak tuning. Values are mean and standard deviation across 24 images. Both methods select their configurations separately for each image.}
\label{tab:staf_per_image_summary}
\begin{tabular}{lcccc}
\toprule
Method & Candidates/image & PSNR (dB) & SSIM & PSNR wins \\
\midrule
LSA & 3 & 38.308 $\pm$ 2.849 & 0.9568 $\pm$ 0.0111 & 24/24 \\
STAF & 56 & 35.425 $\pm$ 2.190 & 0.9209 $\pm$ 0.0119 & 0/24 \\
\bottomrule
\end{tabular}
\end{table}

The mean paired PSNR difference is 2.883 dB, with a 95\% bootstrap interval of $[2.439,3.295]$ dB. LSA also has higher SSIM on every image. These results establish performance under the specified per-image search budgets.

The matched-budget global STAF configuration in \Cref{fig:kodak_bar} is selected from the completed $56\times24$ sweep using mean PSNR on the same six calibration images, then frozen across all 24 images. The selected configuration has $\tau=3$, learning rate $3\times10^{-3}$, and decay to 0.05 of the initial rate over 5,000 updates. Its all-24 mean is 35.369 dB. The extended-budget global STAF result in \Cref{tab:kodak_global} uses a separate historical search over term count, learning rate, and schedule, with up to 10,000 updates and plateau-based early stopping. Its all-24 mean is 35.146 dB. Both full-set means include calibration images and are descriptive.

\paragraph{SL$^2$A.}
We evaluate the official architecture under a shared 5,000-update image-fitting protocol. The full model uses Chebyshev degree 256 and rank 128, with 330,243 parameters. The parameter-matched model uses degree 192 and rank 64, with 199,171 parameters. LSA has 199,043 trainable parameters. Each SL$^2$A variant is calibrated over learning rates $\{5\times10^{-4},10^{-3},2\times10^{-3},4\times10^{-3},8\times10^{-3}\}$ on the same six images. The selected rate is $4\times10^{-3}$ for both full and parameter-matched models, with the same $0.1^{\min(s/5000,1)}$ decay schedule. These settings are frozen before the 18-image evaluation. We also report an author-style SL$^2$A run with degree 256, rank 128, 501 epochs, batch size 16,384, and seed 42. This optimization protocol is reported separately in \Cref{tab:sl2a_kodak}.

\begin{table}[htbp]
\centering
\caption{SL$^2$A comparison on Kodak. The 18-image holdout is the primary comparison after calibration. The 24-image means also include the six calibration images. Wins refer to LSA relative to each row on the holdout.}
\label{tab:sl2a_kodak}
\begin{tabular}{lrrrr}
\toprule
Method & Parameters & Holdout PSNR & All-24 PSNR & LSA wins \\
\midrule
LSA & 199,043 & 38.045 & 38.110 & - \\
SL$^2$A, full & 330,243 & 37.857 & 37.298 & 10/18 \\
SL$^2$A, parameter matched & 199,171 & 35.910 & 35.995 & 18/18 \\
SL$^2$A, author protocol & 330,243 & 37.463 & 37.541 & 15/18 \\
\bottomrule
\end{tabular}
\end{table}

The mean difference for LSA minus full SL$^2$A is $0.188$ dB, with ten wins and eight losses for LSA. The exact sign test and Wilcoxon test give $p=0.815$ and $p=0.229$, respectively. Full SL$^2$A has a slightly lower descriptive mean and uses 65.9\% more parameters. With nearly matched parameter counts, LSA gains 2.135 dB and wins on all 18 images. Against the author-style protocol, LSA gains 0.582 dB with 15/18 wins. These comparisons quantify reconstruction quality and parameter cost under the specified protocols.

\paragraph{WIRE, MFN, and fJNB.}
These comparisons use the same six-image calibration and 18-image holdout. WIRE is selected from 32 candidates covering frequency, spatial scale, learning rate, initialization, and schedule. MFN GaborNet is selected from 36 candidates covering input scale, Gabor scale, and learning rate. Both receive up to 10,000 updates, and LSA receives three calibration candidates and 5,000 updates. The fJNB sweep uses degrees $q\in\{2,3,4,5,6\}$, shared Adam rates $\{10^{-3},3\times10^{-3},10^{-2},3\times10^{-2}\}$, and constant or decayed schedules, giving 240 image-configuration attempts. The selected stable configuration uses $q=5$, learning rate $3\times10^{-3}$, and a constant schedule. This single-degree sequential construction is the scope of the fJNB comparison.

\begin{table}[htbp]
\centering
\caption{Additional Kodak holdout results after six-image calibration. Confidence intervals are paired bootstrap intervals for LSA's mean PSNR gain. Means and paired gains are rounded independently.}
\label{tab:kodak_holdout_extra}
\begin{tabular}{lrrrr}
\toprule
Method & Holdout PSNR & LSA gain (dB) & 95\% interval & LSA wins \\
\midrule
LSA & 38.045 & - & - & - \\
fJNB & 22.985 & 15.060 & [14.595, 15.559] & 18/18 \\
WIRE & 36.477 & 1.568 & [1.143, 1.970] & 17/18 \\
MFN GaborNet & 35.086 & 2.959 & [2.764, 3.157] & 18/18 \\
\bottomrule
\end{tabular}
\end{table}

On the holdout, LSA and fJNB obtain SSIM $0.9545\pm0.0115$ and $0.5449\pm0.1274$, respectively. Their PSNR standard deviations are 2.631 and 2.325 dB. \Cref{tab:kodak_global} collects the descriptive 24-image means for globally selected configurations. These full-set results provide context alongside the holdout and per-image evaluations.

\begin{table}[htbp]
\centering
\caption{Descriptive Kodak results over all 24 images with globally selected configurations. Calibration images are included. STAF, WIRE, and MFN receive up to 10,000 updates. LSA, SIREN, and FINER use 5,000.}
\label{tab:kodak_global}
\begin{tabular}{lr}
\toprule
Method & Mean PSNR (dB) \\
\midrule
SIREN & 34.990 \\
FINER & 37.317 \\
STAF & 35.146 \\
WIRE & 36.596 \\
MFN GaborNet & 35.151 \\
fJNB & 22.948 \\
LSA & 38.110 \\
\bottomrule
\end{tabular}
\end{table}

\begin{table}[htbp]
\centering
\caption{Per-image STAF comparison after selecting from 56 STAF and three LSA configurations per image.}
\label{tab:staf_per_image}
\begin{tabular}{lrrrr}
\toprule
Image & LSA PSNR & STAF PSNR & LSA SSIM & STAF SSIM \\
\midrule
kodim01 & 35.671 & 32.973 & 0.9562 & 0.9240 \\
kodim02 & 39.563 & 35.636 & 0.9529 & 0.8902 \\
kodim03 & 42.714 & 38.166 & 0.9737 & 0.9321 \\
kodim04 & 39.194 & 35.999 & 0.9563 & 0.9086 \\
kodim05 & 35.157 & 31.669 & 0.9605 & 0.9157 \\
kodim06 & 38.192 & 34.668 & 0.9614 & 0.9256 \\
kodim07 & 41.622 & 37.352 & 0.9780 & 0.9388 \\
kodim08 & 32.677 & 31.542 & 0.9368 & 0.9212 \\
kodim09 & 41.458 & 37.500 & 0.9617 & 0.9221 \\
kodim10 & 40.621 & 36.995 & 0.9631 & 0.9218 \\
kodim11 & 38.268 & 35.267 & 0.9570 & 0.9142 \\
kodim12 & 40.604 & 37.835 & 0.9620 & 0.9306 \\
kodim13 & 31.910 & 31.334 & 0.9364 & 0.9323 \\
kodim14 & 36.461 & 33.264 & 0.9568 & 0.9122 \\
kodim15 & 37.810 & 36.565 & 0.9417 & 0.9149 \\
kodim16 & 40.987 & 36.891 & 0.9731 & 0.9268 \\
kodim17 & 40.085 & 37.235 & 0.9664 & 0.9359 \\
kodim18 & 35.968 & 33.370 & 0.9442 & 0.9068 \\
kodim19 & 37.853 & 35.374 & 0.9508 & 0.9190 \\
kodim20 & 39.199 & 37.962 & 0.9572 & 0.9405 \\
kodim21 & 38.674 & 35.635 & 0.9569 & 0.9150 \\
kodim22 & 37.464 & 34.624 & 0.9476 & 0.9024 \\
kodim23 & 41.965 & 38.376 & 0.9675 & 0.9295 \\
kodim24 & 35.286 & 33.968 & 0.9440 & 0.9203 \\
\bottomrule
\end{tabular}
\end{table}

\subsection{Audio Fitting}
\label{app:baseline_audio}

We use the fixed 10-clip NSynth and 40-clip LibriSpeech manifests, one-second mono signals resampled to 48 kHz, waveform MSE, and PSNR with unit dynamic range, as in \Cref{app:impaudio}. The coordinate MLP comparisons use width 256, four hidden layers, Adam, and 5,000 updates. PSNR is selected from evaluated checkpoints on the fitting grid. FF16 denotes 16 Gaussian Fourier-feature bands with the raw coordinate included. Settings are selected separately for each method and evaluated on the same clips. These are signal-fitting comparisons.

\paragraph{STAF.}
We evaluate the official audio-style raw-coordinate configuration with $\tau=5$ and first-layer $\omega_0=3000$, as well as matched Fourier-feature settings at $\sigma=20$ and 100. A $\tau=16$ setting matches LSA's harmonic count. \Cref{tab:audio_staf_sweep} gives all four settings. The raw-coordinate setting is strongest among those tested, reaching 68.79 dB on NSynth and 68.40 dB on LibriSpeech. FINER reaches 67.19 and 66.32 dB, respectively. LSA at FF16, $\sigma=100$, and $K=16$ reaches $86.72\pm21.18$ and $90.78\pm18.57$ dB, with 8/10 and 37/40 wins over the strongest STAF setting.

\begin{table}[htbp]
\centering
\caption{STAF audio configurations. Entries are mean PSNR and standard deviation across clips. The raw-coordinate setting uses first-layer $\omega_0=3000$.}
\label{tab:audio_staf_sweep}
\begin{tabular}{llrr}
\toprule
Input & Terms $\tau$ & NSynth & LibriSpeech \\
\midrule
Raw coordinates & 5 & $68.79\pm3.04$ & $68.40\pm6.29$ \\
FF16, $\sigma=20$ & 5 & $55.11\pm11.63$ & $50.24\pm1.38$ \\
FF16, $\sigma=100$ & 5 & $51.83\pm4.47$ & $51.85\pm1.28$ \\
FF16, $\sigma=100$ & 16 & $50.60\pm2.44$ & $51.21\pm1.43$ \\
\bottomrule
\end{tabular}
\end{table}

\paragraph{SL$^2$A.}
The sweep covers Chebyshev degrees 256 and 512 and ranks 64 and 128, using the official learnable-activation and fusion architecture. \Cref{tab:audio_sl2a} reports the configurations at learning rate $10^{-3}$. Degree 512 and rank 128 give the strongest mean result on both datasets, with 68.03 dB on NSynth and 47.42 dB on LibriSpeech. LSA wins on 7/10 and 40/40 clips, respectively.

\begin{table}[htbp]
\centering
\caption{SL$^2$A audio results for the tested degree and rank settings. Values are mean PSNR and standard deviation across clips. All rows use learning rate $10^{-3}$.}
\label{tab:audio_sl2a}
\begin{tabular}{rrrrr}
\toprule
Degree & Rank & Parameters & NSynth & LibriSpeech \\
\midrule
512 & 128 & 329,473 & $68.03\pm10.20$ & $47.42\pm6.22$ \\
512 & 64 & 231,169 & $67.16\pm10.57$ & $46.35\pm6.62$ \\
256 & 128 & 263,937 & $59.16\pm16.05$ & $38.84\pm7.18$ \\
256 & 64 & 165,633 & $59.47\pm15.35$ & $38.37\pm7.28$ \\
\bottomrule
\end{tabular}
\end{table}

\paragraph{WIRE and fJNB.}
The selected WIRE settings are $(\omega_0,s_0)=(10,10)$ for NSynth and $(10,5)$ for LibriSpeech, both at learning rate $10^{-3}$. For fJNB, the sweep covers degrees $q\in\{2,3,4,5,6\}$, learning rates $\{10^{-3},5\times10^{-4},2\times10^{-4},10^{-4}\}$, and Fourier scales $\sigma\in\{20,100\}$, together with raw-coordinate runs. We report the strongest completed configurations. \Cref{tab:audio_other} includes both input regimes for fJNB and the corresponding LSA results.

\begin{table}[htbp]
\centering
\caption{Additional audio fitting comparisons. Gains and wins refer to LSA relative to the specified baseline on the same clips. fJNB uses the fractional Jacobi construction.}
\label{tab:audio_other}
\begin{tabular}{lllrrrr}
\toprule
Dataset & Baseline & Input & Baseline PSNR & LSA PSNR & Gain & Wins \\
\midrule
NSynth & fJNB & Raw & 16.45 & 47.41 & 30.96 & 9/10 \\
LibriSpeech & fJNB & Raw & 25.23 & 31.96 & 6.73 & 31/40 \\
NSynth & fJNB & FF16, $\sigma=100$ & 55.09 & 86.72 & 31.63 & 10/10 \\
LibriSpeech & fJNB & FF16, $\sigma=100$ & 43.97 & 90.78 & 46.81 & 40/40 \\
NSynth & WIRE & Selected & 47.11 & 86.72 & 39.61 & 10/10 \\
LibriSpeech & WIRE & Selected & 51.79 & 90.78 & 38.99 & 40/40 \\
\bottomrule
\end{tabular}
\end{table}

\subsection{Novel View Synthesis}
\label{app:baseline_nerf}

We adapt the official fractional Jacobi block as an activation in the NeRF backbone. Degree and learning rate are tuned on Chair, Ficus, and Ship using 50,000 steps and mean final validation PSNR, followed by one 150,000-step run on each of the eight scenes. The fJNB and LSA comparison uses the same Blender split, reduced-view protocol, backbone, ray budget, sample count, width, depth, optimizer family, and final-step evaluation rule. This protocol concerns the adapted fJNB block.

\Cref{tab:nerf_extra} also reproduces WIRE PSNR from the reduced-view Blender table in FINER~\citep{liu2024finer}, which uses 25 training views. WIRE is a published reference with its source protocol, while the fJNB and LSA columns are evaluated in the shared backbone.

\begin{table}[htbp]
\centering
\caption{Additional Blender PSNR results. WIRE is reproduced from~\citet{liu2024finer}. fJNB and LSA use the shared-backbone evaluation described above. Means are across eight scenes.}
\label{tab:nerf_extra}
\begin{tabular}{lrrr}
\toprule
Scene & WIRE (published) & fJNB (adapted) & LSA \\
\midrule
Chair & 29.31 & 30.15 & 34.18 \\
Drums & 22.22 & 18.08 & 24.79 \\
Ficus & 25.91 & 22.24 & 28.85 \\
Hotdog & 30.11 & 31.81 & 33.36 \\
Lego & 25.76 & 22.67 & 29.39 \\
Materials & 25.05 & 24.73 & 27.22 \\
Mic & 32.35 & 31.58 & 33.91 \\
Ship & 21.15 & 19.62 & 22.60 \\
\midrule
Mean & 26.48 & 25.11 & 29.29 \\
\bottomrule
\end{tabular}
\end{table}

\subsection{Ablations}
\label{app:lsa_ablations}

The truncation ablation in \Cref{tab:k_ablation} evaluates six Kodak images. Quality increases through $K=32$ and saturates at $K=64$ under this protocol. Together with the runtime and memory measurements in \Cref{tab:resource_k_scaling}, the results identify $K$ as a control on quality and activation cost. The quality and resource tables concern their respective image and audio settings.

\begin{table}[htbp]
\centering
\caption{Harmonic-count ablation on six Kodak images. Standard deviations and medians are computed across images.}
\label{tab:k_ablation}
\begin{tabular}{rrrrr}
\toprule
$K$ & Mean PSNR & Std. PSNR & Median PSNR & Mean SSIM \\
\midrule
4 & 30.44899 & 5.28322 & 29.17142 & 0.81556 \\
8 & 35.35890 & 4.63919 & 34.52907 & 0.92290 \\
16 & 38.67377 & 4.34150 & 40.33058 & 0.96070 \\
32 & 39.50432 & 4.32379 & 41.38003 & 0.96614 \\
64 & 39.34945 & 4.14439 & 40.49803 & 0.96494 \\
\bottomrule
\end{tabular}
\end{table}

For the frozen-coefficient ablation, both variants use raw coordinates, $K=32$, the same initial state, data order, random seed, optimizer settings, and 5,000-step budget. The frozen variant keeps $a$ at its small-random initialization while training all ordinary weights and biases. The coefficients were verified to remain bitwise unchanged. Hyperparameters are selected on six calibration images, with the other 18 used for the primary evaluation.

\begin{table}[htbp]
\centering
\caption{Learning versus freezing the near-identity coefficients. PSNR is mean and standard deviation across images. Confidence intervals refer to the paired mean PSNR gain.}
\label{tab:frozen_coefficients}
\begin{tabular}{llrrrr}
\toprule
Evaluation & Coefficients & PSNR & SSIM & LSA wins & Gain 95\% interval \\
\midrule
18 holdout & Learned & $38.05\pm2.63$ & 0.9545 & 18/18 & $[17.45,18.87]$ \\
18 holdout & Frozen & $19.91\pm1.78$ & 0.4861 & & \\
24 pooled & Learned & $38.11\pm2.90$ & 0.9557 & 24/24 & $[17.56,19.17]$ \\
24 pooled & Frozen & $19.76\pm1.95$ & 0.4834 & & \\
\bottomrule
\end{tabular}
\end{table}

The paired mean PSNR gains are 18.13 dB on the holdout and 18.35 dB on all 24 images. The frozen activation remains close to the identity, so this comparison measures the contribution of learning spectral coefficients from that initialization. It establishes that the learned nonlinear response is important for the reported fits. The result is specific to this near-identity frozen control.

\section{Additional Analysis}
\label{app:addan}
In this section we provide additional analysis of our proposed method and its shaping of the INR.
\subsection{Hidden-Representation Diagnostics}
\label{app:hidden_diagnostics}

\begin{figure}[t]
    \centering
    \includegraphics[width=\linewidth]{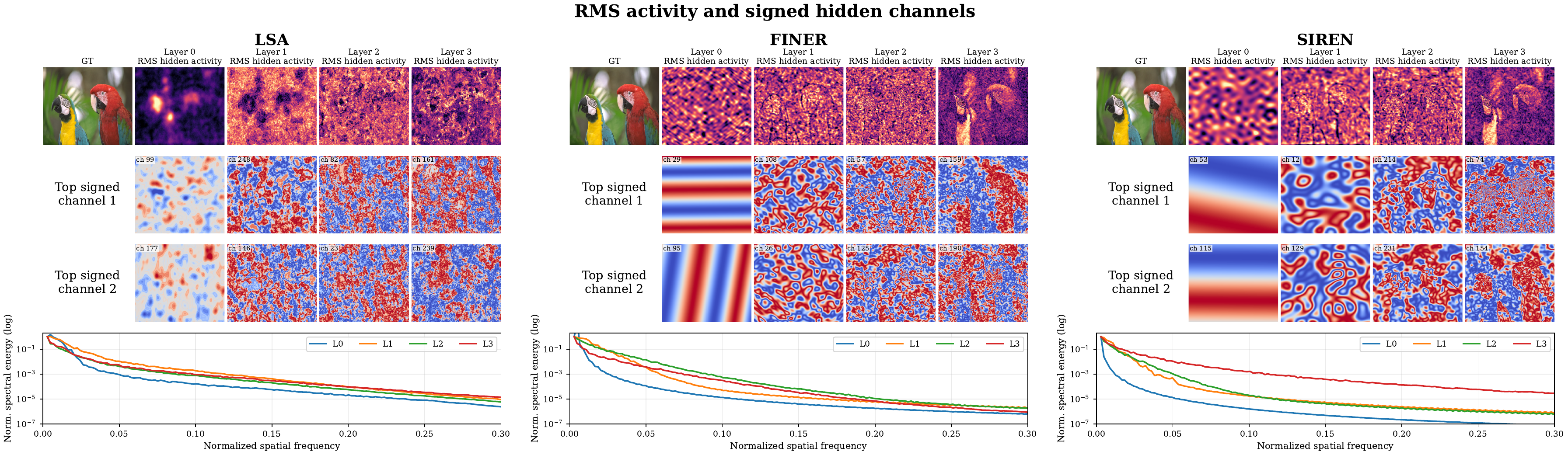}
    \caption{
    Hidden-representation diagnostics on image fitting. For each method, we visualize layer-wise RMS hidden activity, the two highest-variance signed hidden channels, and the radial spectra of those selected channels. LSA produces more spatially localized and less purely periodic hidden features than SIREN and FINER, supporting the view that learned spectral activations modify the internal basis rather than only the final reconstruction.
    }
    \label{fig:hidden_diagnostics}
\end{figure}

To further inspect how learned spectral activations modify the internal representation, we visualize hidden features from trained image INRs in \Cref{fig:hidden_diagnostics}. For each method and layer, we evaluate the network on a dense coordinate grid and record the hidden activations $h^{(\ell)}(x)\in\mathbb{R}^{C}$. We summarize the layer-wise activity by the RMS map
\[
A^{(\ell)}(x)
=
\left(
\frac{1}{C}\sum_{c=1}^{C}
\left(h^{(\ell)}_{c}(x)\right)^2
\right)^{1/2},
\]
which indicates where hidden units are active after aggregating over channels. We also show the two channels with largest spatial variance in each layer, preserving their sign, to visualize representative high-energy hidden basis functions.

The diagnostic reveals a qualitative difference in how the methods organize their hidden representations. SIREN and FINER exhibit more globally periodic early-layer channels, consistent with fixed sinusoidal or frequency-warped nonlinear operators. Their later layers gradually become more image-aligned, suggesting that localized structure is assembled through depth. In contrast, LSA produces less purely periodic and more spatially localized hidden features already in early layers. This supports the mechanism proposed in \Cref{sec:method}: learned spectral activations change the neuron-level basis used to compose heterogeneous structure, rather than only improving the final reconstruction through the output layer.

The bottom row of \Cref{fig:hidden_diagnostics} shows radial spectra of the selected high-variance channels. These spectra should be interpreted as a diagnostic of dominant hidden components rather than a complete spectrum of the full representation, since they are computed only from the selected channels. Nevertheless, they are consistent with the spatial visualizations: LSA exhibits similar spectral behavior across layers, while SIREN and FINER show stronger layer-wise changes. This suggests that LSA expresses part of the required spectral complexity directly inside individual hidden responses, whereas fixed activation families rely more heavily on progressive composition across layers.

\subsection{Learned Activation Coefficient Spectra}
\label{app:coefficient_spectra}

In addition to visualizing hidden activations, we inspect the learned spectral operators themselves. For each trained image INR, LSA contains layer-wise activation coefficients $\{a_k^{(\ell)}\}_{k=1}^{K}$ in \Cref{eq:method_activation}. These coefficients are model parameters, not data-dependent activations. We therefore visualize them directly to determine whether the learned activation uses a nontrivial harmonic profile.

\begin{figure}[t]
    \centering
    \includegraphics[width=\linewidth]{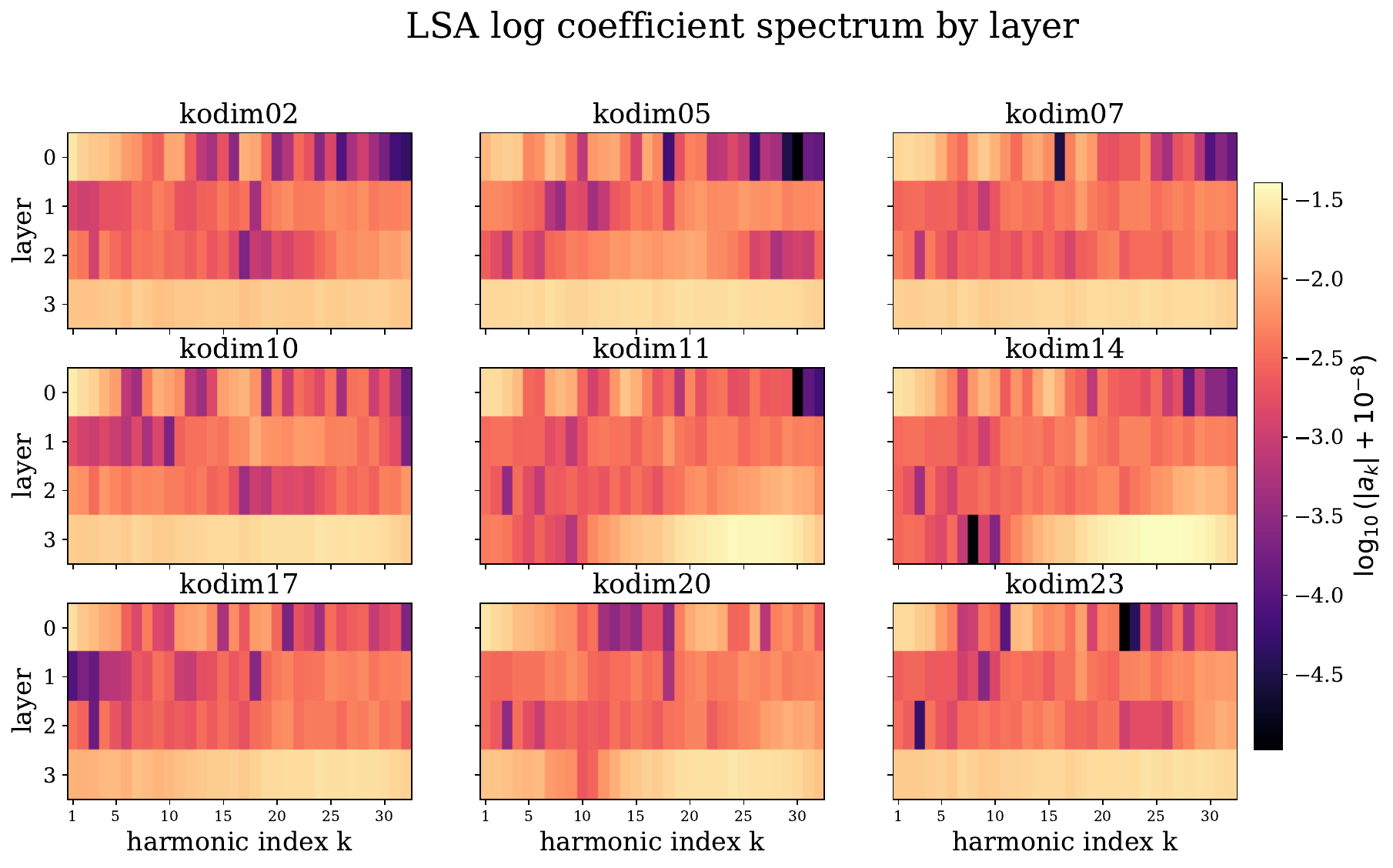}
    \caption{
    Learned LSA coefficient spectra across Kodak images. Each panel shows $\log_{10}(|a_k^{(\ell)}|+10^{-8})$ for harmonic index $k$ and layer $\ell$. The learned operators exhibit layer-dependent harmonic profiles and image-dependent variation, indicating that LSA adapts its internal spectral basis rather than using a fixed spectrum.
    }
    \label{fig:lsa_coeff_spectra}
\end{figure}

For each layer $\ell$, we plot
\[
M_{\ell,k}
=
\log_{10}\left(|a_k^{(\ell)}| + 10^{-8}\right),
\]
where $k$ indexes the harmonic component. Each heatmap in \Cref{fig:lsa_coeff_spectra} corresponds to a separately trained INR on one Kodak image. Rows denote layers and columns denote harmonic indices.

The learned coefficients exhibit clear layer-dependent structure. Early layers often show selective harmonic suppression or enhancement, visible as dark or bright bands across specific harmonic indices. Later layers, especially the final hidden layer, tend to retain broader and more uniform harmonic support. The overall pattern is consistent across images, while the exact emphasized or suppressed harmonics vary, indicating that the learned spectral operator adapts to the fitted signal rather than collapsing to a fixed handcrafted profile.

This diagnostic is complementary to the hidden-activation visualization in \Cref{fig:hidden_diagnostics}. The coefficient spectra show that LSA learns nontrivial harmonic operators in parameter space, while the hidden-channel diagnostics show how these operators manifest as spatial features over the coordinate domain.

\subsection{Learned Activation Profiles}
\label{app:learned_activation_profiles}

We further inspect the activation functions learned by LSA on Kodak image fitting. For each trained image INR, the activation at layer $\ell$ is parameterized by the coefficients $\{a_k^{(\ell)}\}_{k=1}^{K}$ in \Cref{eq:method_activation}. We visualize these parameters in two complementary ways: first, by plotting the coefficient magnitudes $|a_k^{(\ell)}|$ as a function of harmonic index $k$, and second, by plotting the resulting learned activation waveform $\phi^{(\ell)}(u)$ over the pre-activation range $u\in[-1,1]$.

\begin{figure}[t]
    \centering
    \includegraphics[width=\linewidth]{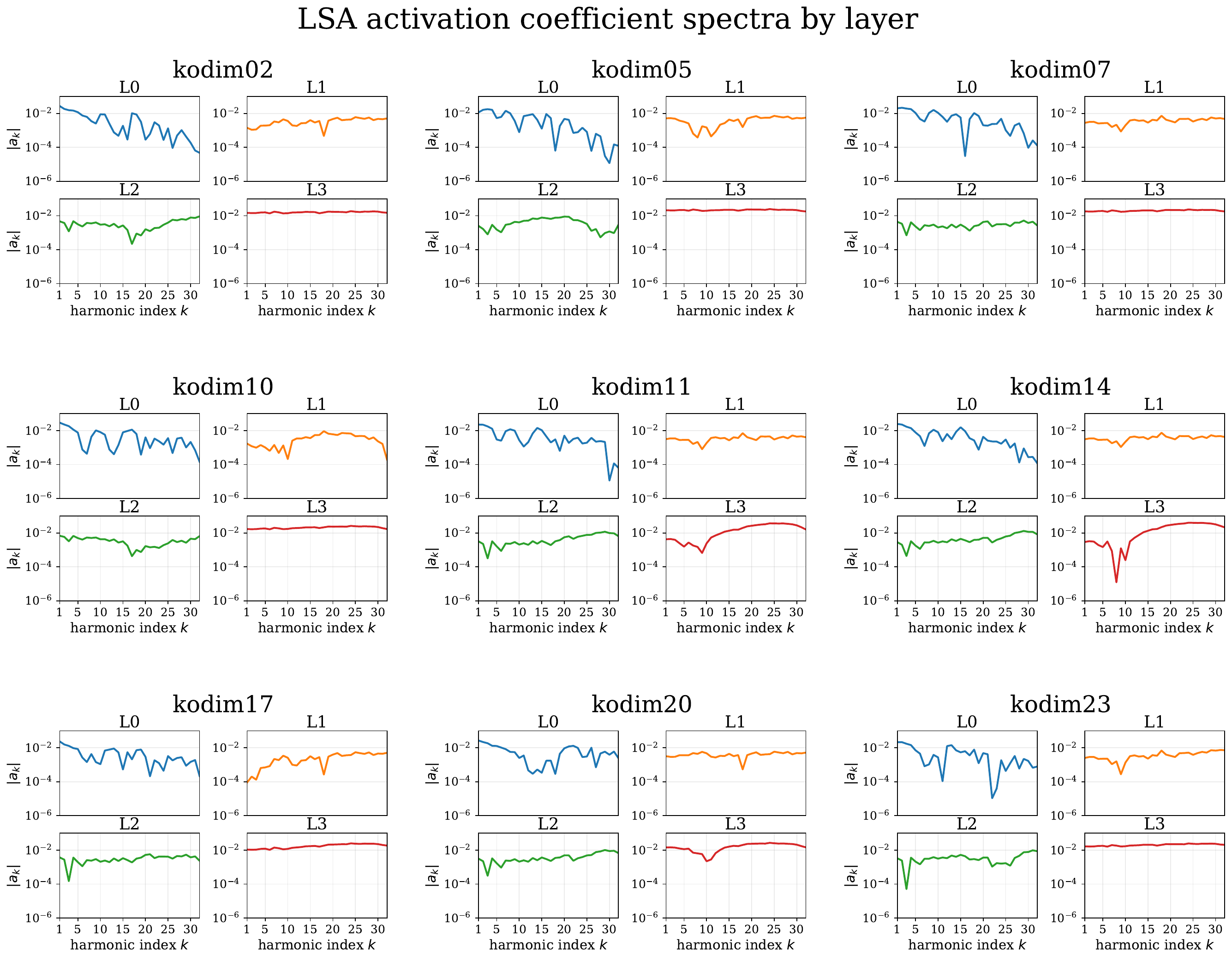}
    \caption{
    Learned activation coefficient spectra across Kodak images. Each panel shows the magnitude $|a_k^{(\ell)}|$ of the learned harmonic coefficients for one image and layer. The coefficients exhibit layer-dependent profiles and image-dependent variation, indicating that LSA learns nontrivial spectral operators rather than using a fixed harmonic profile.
    }
    \label{fig:lsa_coeff_spectra_grid}
\end{figure}

\begin{figure}[t]
    \centering
    \includegraphics[width=\linewidth]{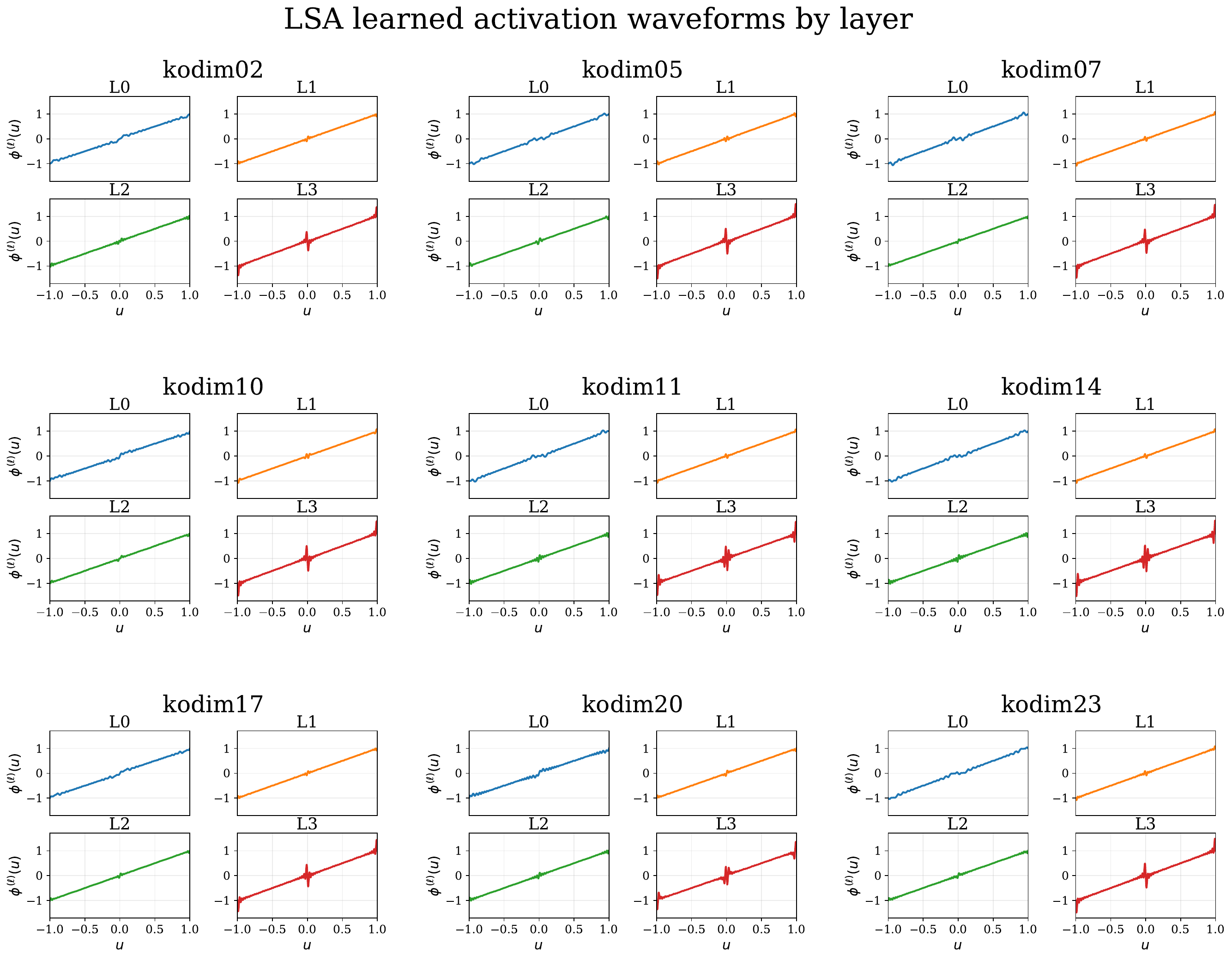}
    \caption{
    Learned activation waveforms across Kodak images. Each panel plots the resulting activation $\phi^{(\ell)}(u)$ for one image and layer. Although the learned functions remain close to the residual identity shape, they contain localized harmonic deviations whose form varies across layers and fitted images.
    }
    \label{fig:lsa_activation_waveforms}
\end{figure}

The coefficient spectra in \Cref{fig:lsa_coeff_spectra} show that the learned harmonic profiles are structured rather than uniform. Early layers often display selective attenuation or enhancement of specific harmonics, while later layers tend to retain broader harmonic support. The precise profile varies across images, suggesting that the learned operator adapts to the fitted signal.

The waveform plots in \Cref{fig:lsa_activation_waveforms} provide the corresponding activation-space view. The learned activations remain close to the residual identity mapping, which is consistent with the small initialization used in \Cref{sec:method}. At the same time, the curves contain localized deviations and small oscillatory corrections, especially around frequently used pre-activation regions. This supports the intended role of LSA: the method does not require large departures from a stable base activation, but learns targeted harmonic corrections that reshape the hidden response.

Together, these diagnostics complement the hidden-representation analysis in \Cref{app:hidden_diagnostics}. The coefficient plots show that the spectral operator itself is learned in parameter space, while the waveform plots show how these coefficients modify the pointwise nonlinear response applied inside the INR.

\subsection{Resource Analysis}
\label{app:resource_analysis}

LSA is compact in trainable parameters, but its current implementation introduces additional activation compute. When coefficients are shared within each activation layer, LSA adds only \(K L_{\mathrm{act}}\) trainable scalar parameters, where \(K\) is the number of harmonic terms and \(L_{\mathrm{act}}\) is the number of activation layers. For example, in our reported width-256, depth-4 audio INR, \(K=16\) adds only \(16 \times 4 = 64\) trainable parameters.

The main overhead comes from evaluating the harmonic expansion pointwise at every hidden activation. For a batch of \(B\) coordinates, hidden width \(W\), \(L_{\mathrm{act}}\) activation layers, and \(K\) harmonics, the activation evaluates \(B W L_{\mathrm{act}} K\) harmonic responses. Thus, the practical cost is dominated by intermediate activation tensors and autograd storage rather than by the number of trainable parameters.

To quantify this cost, we benchmarked SIREN, FINER, and LSA under matched input regimes. In the primary comparison, all methods use the same FF16 input encoding. Runtime is measured on an NVIDIA RTX A6000 using full-grid 48k audio fitting. Training time includes forward pass, backward pass, and optimizer update. Evaluation time is forward-only. Memory is peak CUDA allocation. LSA is measured using the batched PyTorch implementation used in our experiments.

\begin{table}[t]
\centering
\caption{
Resource usage under the matched FF16 input regime on full-grid 48k audio fitting.
All models use width 256. Training time includes forward, backward, and optimizer update.
Evaluation time is forward-only. Memory is peak CUDA allocation.
}
\label{tab:resource_ff16}
\resizebox{\linewidth}{!}{
\begin{tabular}{llrrrrrr}
\toprule
Depth & Method
& Params
& Extra act. params
& Train time (ms)
& Train mem. (GB)
& Eval time (ms)
& Eval mem. (GB) \\
\midrule
\multirow{3}{*}{4}
& SIREN FF16
& 206{,}337 & 0
& 8.26 & 0.48 & 3.56 & 0.16 \\
& FINER FF16
& 206{,}337 & 0
& 14.20 & 0.85 & 5.55 & 0.21 \\
& LSA FF16, \(K=16\)
& 206{,}401 & 64
& 97.93 & 7.53 & 35.08 & 1.58 \\
\midrule
\multirow{3}{*}{6}
& SIREN FF16
& 337{,}921 & 0
& 12.77 & 0.67 & 5.46 & 0.16 \\
& FINER FF16
& 337{,}921 & 0
& 21.65 & 1.22 & 8.45 & 0.21 \\
& LSA FF16, \(K=16\)
& 338{,}017 & 96
& 147.84 & 10.56 & 52.87 & 1.58 \\
\bottomrule
\end{tabular}
}
\end{table}

The results show that LSA has negligible parameter overhead but higher activation cost in the current implementation. At depth 4 and \(K=16\), LSA adds only 64 parameters over the FF16 baselines, while requiring additional runtime and peak memory due to evaluating and storing the harmonic responses. This reflects a tradeoff between a richer learned activation operator and per-step computational cost. Since several experiments also show improved optimization trajectories, including the convergence curves in \Cref{fig:kodak_convergence_full}, part of this per-step cost may be offset in settings where LSA reaches a target reconstruction quality in fewer optimization steps. We leave a systematic quality-vs-wall-clock analysis to future work.

\paragraph{Scaling with the number of harmonics.}
The activation cost scales with the number of harmonic terms \(K\). \Cref{tab:resource_k_scaling} reports LSA resource usage for \(K \in \{8,16,32,64\}\) under the same FF16 full-grid setting.

\begin{table}[t]
\centering
\caption{
LSA scaling with the number of harmonic terms \(K\) under the matched FF16
full-grid 48k setting. All models use width 256. The implementation is the
batched PyTorch harmonic evaluation used in \Cref{tab:resource_ff16}.
}
\label{tab:resource_k_scaling}
\resizebox{\linewidth}{!}{
\begin{tabular}{lrrrrr}
\toprule
Depth & \(K\)
& Train time (ms)
& Train mem. (GB)
& Eval time (ms)
& Eval mem. (GB) \\
\midrule
\multirow{4}{*}{4}
& 8  & 52.78  & 3.88  & 19.24  & 0.85 \\
& 16 & 98.57  & 7.53  & 35.33  & 1.58 \\
& 32 & 189.87 & 14.87 & 67.26  & 3.05 \\
& 64 & 341.09 & 29.55 & 115.23 & 5.98 \\
\midrule
\multirow{4}{*}{6}
& 8  & 79.13  & 5.44  & 28.77  & 0.85 \\
& 16 & 147.84 & 10.56 & 52.87  & 1.58 \\
& 32 & 284.81 & 20.82 & 100.88 & 3.05 \\
& 64 & 511.70 & 41.36 & 172.70 & 5.98 \\
\bottomrule
\end{tabular}
}
\end{table}

As expected, both runtime and memory increase with \(K\), since each additional harmonic introduces another pointwise sinusoidal response. This makes \(K\) a direct knob for trading spectral flexibility against computational cost.

\paragraph{Implementation dependence.}
The measured overhead also depends on how the harmonic activation is implemented. A batched implementation materializes the harmonic tensor over the \(K\) terms, while an accumulate implementation evaluates the terms sequentially:
\[
    \phi(u)
    =
    u + \sum_{k=1}^{K} a_k \sin(2\pi k u).
\]
The batched implementation is faster, but uses more memory. The accumulate implementation reduces memory, especially during inference, but is slower.

\begin{table}[t]
\centering
\caption{
Implementation diagnostic for LSA in a no-FF, small-batch setting
(\(B=4096\), width 256, depth 4, fp32). The accumulate implementation avoids
explicitly materializing the full harmonic tensor, reducing memory at the cost
of additional runtime.
}
\label{tab:resource_impl_diagnostic}
\resizebox{\linewidth}{!}{
\begin{tabular}{llrrrr}
\toprule
\(K\) & Implementation
& Train time (ms)
& Train mem. (GB)
& Eval time (ms)
& Eval mem. (GB) \\
\midrule
8
& Batched
& 6.63 & 0.35 & 2.26 & 0.09 \\
& Accumulate
& 6.98 & 0.30 & 2.35 & 0.04 \\
\midrule
16
& Batched
& 11.43 & 0.66 & 4.08 & 0.15 \\
& Accumulate
& 13.13 & 0.55 & 4.40 & 0.04 \\
\bottomrule
\end{tabular}
}
\end{table}

This diagnostic suggests that the overhead is partly implementation-dependent. In particular, inference memory can be reduced by avoiding materialization of the full harmonic tensor. More optimized implementations, such as fused CUDA kernels, checkpointed harmonic evaluation, or adaptive sparsification over active harmonics, could further reduce this cost without changing the mathematical form of LSA. Thus, the current implementation should be viewed as a straightforward reference implementation rather than a fully optimized activation kernel.

\newpage

\end{document}